\documentclass[10pt,twocolumn,letterpaper]{article}

\usepackage[pagenumbers]{cvpr} 

\usepackage{algorithm}
\usepackage[accsupp]{axessibility}
\usepackage{algpseudocode}
\usepackage{bm}
\usepackage{booktabs}
\usepackage{mathtools}

\definecolor{cvprblue}{rgb}{0.21,0.49,0.74}
\usepackage[pagebackref,breaklinks,colorlinks,allcolors=cvprblue]{hyperref}

\title{FlashPortrait: 6$\times$ Faster Infinite Portrait Animation \\ with Adaptive Latent Prediction}
\author{Shuyuan Tu$^{1}$ \ \ 
Yueming Pan$^{3}$ \ \ 
Yinming Huang$^{1}$ \ \ 
Xintong Han$^4$ \ \ 
Zhen Xing$^5$ \ \ \vspace{-0.01cm}\\
Qi Dai$^{2}$ \ \ 
Kai Qiu$^{2}$ \ \ 
Chong Luo$^{2}$ \ \ 
Zuxuan Wu$^1$ \\
{$^1$Fudan University}  \quad  
{$^2$Microsoft Research Asia}  \quad  
{$^3$Xi'an Jiaotong University}  \quad  \\
{$^4$Tencent Inc.} \quad
{$^5$Tongyi Lab, Alibaba Group} \\
{\url{https://francis-rings.github.io/FlashPortrait}}
}

\begin{document}

\twocolumn[{
\maketitle
\vspace{-3.2em}
\renewcommand\twocolumn[1][]{#1}
\begin{center}
    \centering
    \includegraphics[width=1\textwidth]{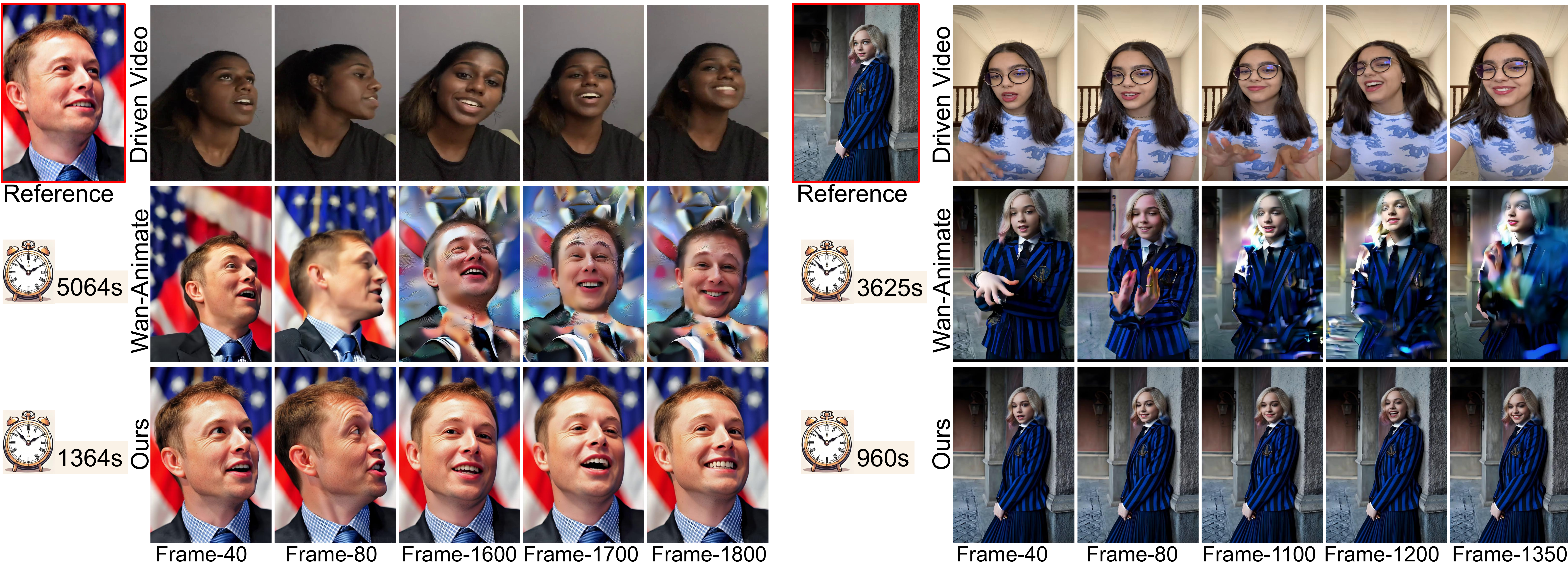}
    \vspace{-0.7cm}
    \captionof{figure}{Portrait animations generated by FlashPortrait, showing its power to synthesize infinite-length ID-preserving animations. 
    Frame-X refers to the X-th frame of the synthesized video.
    The clock icon denotes inference time. 
    Wan-Animate is the latest animation model.
    }
    \label{fig:cover}
\end{center}
}
]

\begin{abstract}
Current diffusion-based acceleration methods for long-portrait animation struggle to ensure identity (ID) consistency. This paper presents FlashPortrait, an end-to-end video diffusion transformer capable of synthesizing ID-preserving, infinite-length videos while achieving up to 6$\times$ acceleration in inference speed. 
In particular, FlashPortrait begins by computing the identity-agnostic facial expression features with an off-the-shelf extractor.
It then introduces a Normalized Facial Expression Block to align facial features with diffusion latents by normalizing them with their respective means and variances, thereby improving identity stability in facial modeling.
During inference, FlashPortrait adopts a dynamic sliding-window scheme with weighted blending in overlapping areas, ensuring smooth transitions and ID consistency in long animations. In each context window, based on the latent variation rate at particular timesteps and the derivative magnitude ratio among diffusion layers, FlashPortrait 
utilizes higher-order latent derivatives at the current timestep to directly predict latents at future timesteps, thereby skipping several denoising steps and achieving 6$\times$ speed acceleration.
Experiments on benchmarks show the effectiveness of FlashPortrait both qualitatively and quantitatively.

\end{abstract}    
\vspace{-0.6cm}

\section{Introduction}
\label{sec:intro}

Portrait animation aims to synthesize portrait videos with natural facial movements, given a reference image and a driven video, with broad applications in film production and virtual assistants.
Diffusion models~\cite{xing2024aid,dhariwal2021diffusion,ho2020denoising,ho2022cascaded,song2020score,song2020denoising,rombach2022high,meng2021sdedit,hertz2022prompt,tumanyan2023plug,weng2024genrec,weng2024art,tu2024motioneditor,tu2024motionfollower,tu2025stableanimator,tu2025stableanimator++} have significantly inspired research in portrait animation~\cite{qiu2025skyreels, xu2025hunyuanportrait, xie2024x_portrait, wang2025fantasyportrait, cheng2025wan_animate}.
To further make portrait animation applicable in real-world scenarios, researchers are advancing toward long-length portrait animation, which substantially raises inference latency and thus drives the need for acceleration strategies. 
However, current acceleration methods fail to sustain coherent portrait animation over extended sequences. Beyond approximately 20 seconds of animation, they commonly suffer from body distortions and identity (ID) inconsistencies, thereby limiting their practical applicability.

To address this issue, some methods have explored quality-preserving acceleration strategies (cache-based methods~\cite{zou2024accelerating_1, zou2024accelerating_2, selvaraju2024fora, liu2025teacache, zheng2025forecast, liu2025reusing} and distillation-based methods~\cite{yin2024one, yin2024improved, huang2025selfforcing, cui2025selfforcing++}) for Image-to-Video (I2V) generation with subtle motion, yet their effectiveness remains limited when applied to long-length portrait animation featuring complex and large-scale facial expressions. 
Cache-based methods use a training-free save-and-reuse approach to skip denoising steps. However, simply reusing previous feature caches for future latents can drift denoising direction, especially in videos with significant motion, as accurately reconstructing future latents with dramatic motion is highly challenging.
By contrast, distillation-based methods require substantial computational cost to train a 4-step student model and rely on autoregressive sampling for long video synthesis. As the student network cannot fully preserve the teacher’s priors, small latent mismatches emerge at every generated segment. These mismatches propagate across segments and intensify over time, ultimately manifesting as noticeable distribution shifts and color instability, especially in long sequences with large and complex motions.
Thus, preserving stable identity in fast, extended-length portrait animations remains challenging.

\begin{figure*}[t!]
\begin{center}
\includegraphics[width=1\linewidth]{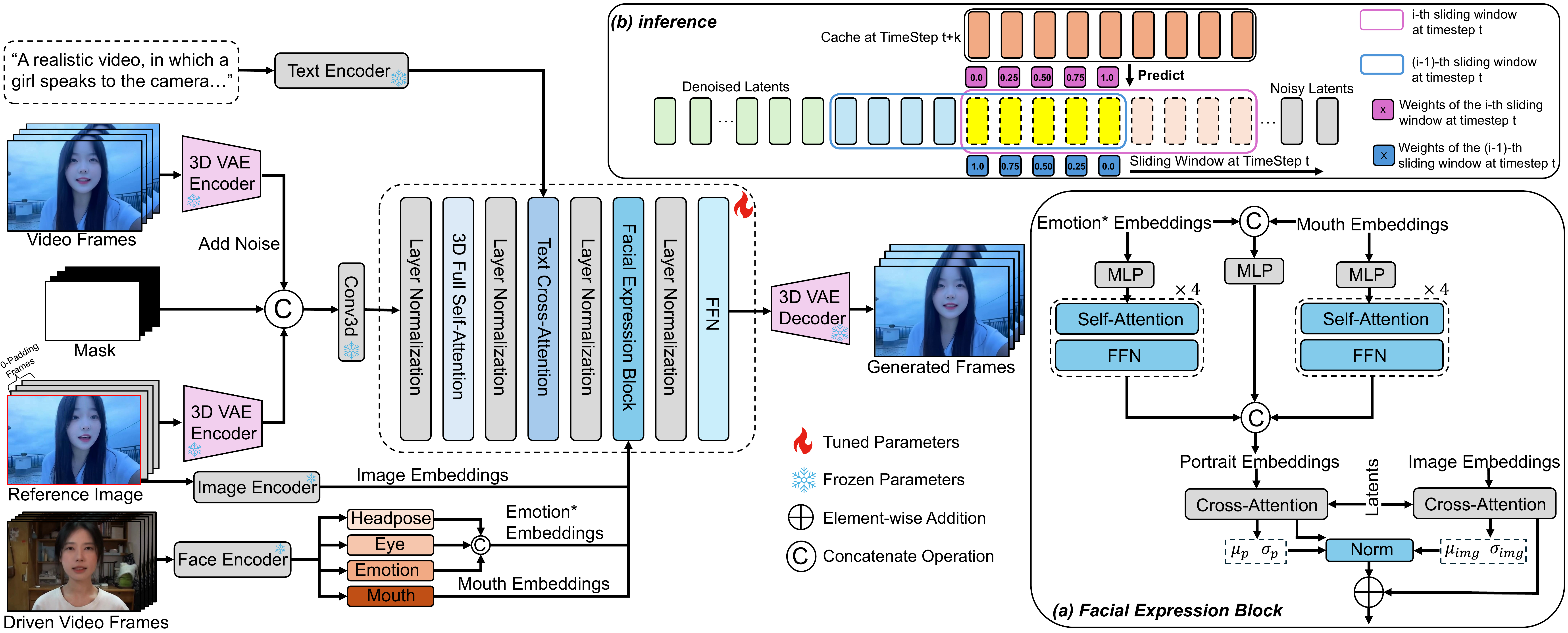}
\end{center}
\vspace{-0.6cm}
   \caption{Architecture of FlashPortrait. (a) and (b) refer to the structure of the Facial Expression Block and long-length video generation pipeline. Embeddings from the Image Encoder and Face Encoder are injected to each block of DiT. 
   To speed up sliding window computation, each window predicts future latents from cached historical states, rather than invoking DiT for denoising.
   }
\label{fig:framework}
\vspace{-0.5cm}
\end{figure*}

Motivated by this, we propose FlashPortrait, a framework with tailored inference and training designs for fast, ID-preserving portrait animation over extended durations, as shown in Fig. \ref{fig:framework}.
To ensure both high-speed acceleration and ID-preserving infinite-length animation, FlashPortrait introduces a novel Sliding Window-based Adaptive Latent Prediction Acceleration Mechanism, which maintains ID consistency while achieving 6$\times$ speedup. 
Concretely, to improve the smoothness of the clip transition in long video generation, FlashPortrait first proposes a weighted sliding-window denoising strategy that fuses latents with progressive weights over time. In each context window, FlashPortrait introduces the Adaptive Latent Prediction Acceleration Mechanism to speed up the window-wise denoising, which leverages the differences among historical latents from previous timesteps to approximate high-order derivatives of the current latent at the present timestep, and then applies the Taylor series to directly predict latents at future timesteps, thereby skipping several denoising steps. 
However, due to the complex and large-amplitude facial motions in portrait animation, latent variations across timesteps are substantial, making fixed-order prediction unreliable~\cite{liu2025reusing, taylor1717methodus, zheng2025forecast}. To address this, based on the latent variation rate at a particular timestep and the derivative magnitude ratio among diffusion layers, we compute two dynamic functions, which adaptively adjust the Taylor expansion, enabling multiple denoising step skipping while ensuring ID stability.

Furthermore, we observe that even within the same clip, ID consistency across frames synthesized by previous models is unstable, primarily due to the large distance between the distribution centers of diffusion latents and facial expression features. To tackle this, FlashPortrait plugs Normalized Facial Expression Blocks into a Video Diffusion Transformer, which significantly enhances ID stability across frames. 
In particular, FlashPortrait first utilizes an off-the-shelf extractor~\cite{wang2022pdfgc} to obtain facial expression features, which are then passed through several self-attention blocks to enhance the perception of the overall facial layout. Then, the means and variances of both the processed features and diffusion latents are computed. FlashPortrait normalizes the facial expression features using these statistics, thereby substantially reducing the adverse effects caused by the large distance between these two distribution centers (latents and facial expression), thereby improving ID stability.

As shown in Fig. \ref{fig:cover} and Table \ref{table:ablation_acceleration}, while the latest open-source portrait animation model Wan-Animate~\cite{cheng2025wan_animate} suffers from dramatic identity inconsistency, color drift, and time-consuming inference, FlashPortrait accurately manipulates the reference based on the driven video while remaining identity stable, achieving a 6$\times$ inference speedup compared with the baseline even when synthesizing extended sequences exceeding 1,800 frames.

In conclusion, our contributions are as follows:
(1) We propose a novel Sliding Window-based Adaptive Latent Prediction Acceleration Mechanism. It is training-free and only activated during inference, achieving a 6$\times$ speedup while maintaining identity consistency in infinite-length portrait animation. To our knowledge, we are the first to explore video diffusion for accelerating ID-preserving infinite-length portrait animation.
(2) We propose a novel Normalized Facial Expression Block to align the distribution centers of diffusion latents and facial features, thereby enhancing identity stability during denoising.
(3) Experimental results on benchmark datasets show the superiority of our model over the SOTA.
\section{Related Work}
\label{sec:related_work}

\noindent\textbf{Video Generation.}
The superior diversity and high fidelity in diffusion models~\cite{dhariwal2021diffusion,ho2020denoising,ho2022cascaded, tu2023implicit,nichol2021improved,song2020score,song2020denoising,rombach2022high,meng2021sdedit, tu2025stableanimator++, tu2025stableavatar, tu2022multiple, tu2023implicit, xing2024survey} has facilitated the advancement of video generation. Early video diffusion works~\cite{singer2022make, blattmann2023stable, guo2023animatediff, tu2024motioneditor, videoworldsimulators2024,tu2024motionfollower,tu2025stableanimator,xing2024simda} mostly are based on the U-Net architecture for video generation by inserting additional temporal layers to pretrained image diffusion models.
Recent works~\cite{bao2024vidu, hong2022cogvideo, kong2024hunyuanvideo, wan2025} replace the U-Net with the Diffusion-in-Transformer (DiT)~\cite{peebles2023scalable} for scalability and higher resolution. Inspired by previous works~\cite{wang2025fantasyportrait, cheng2025wan_animate}, we utilize Wan2.1~\cite{wan2025} as the backbone. 

\noindent\textbf{Portrait Animation.}
It aims to transfer facial motion from a given video to a reference image. Early works~\cite{guo2024liveportrait, hong2023implicit, hong2022depth, li2023g2l, wang2022one, xu2024vasa, zhang2023metaportrait, zhao2022thin} basically apply GANs~\cite{goodfellow2020generative} to model the motion dynamics.
Recently, some studies have applied diffusion models to this field.
FollowYE~\cite{ma2024follow} uses facial keypoints, and Skyreels-A1~\cite{qiu2025skyreels} applies 3D Morphable Models to model facial motion. FantasyPortrait~\cite{wang2025fantasyportrait} supports multi-character portrait animation, and Wan-Animate~\cite{cheng2025wan_animate} unifies portrait animation and character replacement. However, prior DiT-based approaches entail high inference latency and exhibit identity inconsistency and color drift when generating long videos. FlashPortrait addresses these issues and performs ID-preserving infinite-length portrait animation with a 6$\times$ faster inference speed.

\noindent\textbf{Acceleration.}
Acceleration techniques for diffusion models can be broadly categorized as training-free cache-based methods~\cite{zou2024accelerating_1, zou2024accelerating_2, selvaraju2024fora, liu2025teacache, zheng2025forecast} and training-intensive distillation-based methods~\cite{yin2024one, yin2024improved, huang2025selfforcing, cui2025selfforcing++}. Regarding cache-based methods, 
FORA~\cite{selvaraju2024fora} reuses historical attention and MLP features.
TeaCache~\cite{liu2025teacache} decides the caching latents based on the timestep difference estimation. FoCa~\cite{zheng2025forecast} treats feature caching as an ODE solving problem. In terms of distillation-based methods, Self-Forcing~\cite{huang2025selfforcing} and Self-Forcing++~\cite{cui2025selfforcing++} both require extensive GPU resources to train a 4-step student model and generate long videos through an autoregressive sampling paradigm. 
However, the above techniques mainly work for I2V generation tasks with minor motion variations. When applied to long-length portrait animation with large facial motions, the variance of latent changes across timesteps becomes significant, causing the accumulated errors to grow rapidly over time, which eventually leads to ID inconsistency and color drift. 
By contrast, FlashPortrait attains a 6$\times$ inference speed-up while preserving identity consistency and avoiding color drift.

\section{Method}
\label{sec:method}

Illustrated in Fig. \ref{fig:framework}, FlashPortrait builds on Wan2.1~\cite{wan2025} and synthesizes infinite-length ID-preserving animations at a fast speed. 
In particular, the driven video is first fed to PD-FGC~\cite{wang2022pdfgc} to obtain raw facial embeddings (head pose, eyes, emotion, and mouth), which are subsequently refined to reduce the distribution gap between diffusion latents and facial embeddings. 
More details are described in Sec. \ref{sec: face_block}.
Following~\cite{wan2025}, a reference image is incorporated via two pathways. First, the reference is processed by an image encoder~\cite{radford2021learning} to gain image embeddings. These embeddings are injected into Facial Expression Blocks, modulating facial attributes. 
Second, the reference is temporally padded with zero frames and encoded by a frozen 3D VAE~\cite{wan2025} to obtain a latent code. The code is then concatenated channel-wise with compressed video frames and a binary mask (the first frame is 1 and all subsequent frames are 0).

During inference, the original video frames are replaced with random noise, while rest inputs remain unchanged. We further propose a novel Sliding Window-based Adaptive Latent Prediction Acceleration Mechanism, which achieves 6$\times$ inference speedup for ID-preserving infinite-length portrait animation, as detailed in Sec. \ref{sec: sling_window} and Sec. \ref{sec: acceleration}.

\subsection{Normalized Facial Expression Block}
\label{sec: face_block}
We observe that identity consistency across frames synthesized by previous models is unstable, even in the same video clip. The primary limitation lies in their facial modeling, where the large gap between the distribution centers of diffusion latents and raw facial embeddings leads to unstable facial modeling. To address this, we propose a novel Normalized Facial Expression Block to replace each Image Cross-attention block in a denoising DiT.

Concretely, the driven video is fed to a Face Encoder~\cite{wang2022pdfgc} to obtain headpose/eye/emotion and mouth embeddings $\bm{emb}_{m}$, and headpose/eye/emotion are concatenated to gain $\bm{emb}_{e*}$.
The image embeddings $\bm{emb}_{img}$ from the CLIP Image Encoder~\cite{radford2021learning}, $\bm{emb}_{m}$, and $\bm{emb}_{e*}$ are fed to Facial Expression Block to modulate the synthesized identity. We further apply several self-attention $\mathtt{SA}(\cdot)$ and FFN on $\bm{emb}_{m}$ and $\bm{emb}_{e*}$ to enhance the their perception of the overall facial layout, and concatenate the outputs:
\begin{equation}\small
\label{eq:portrait_embeddings}
\begin{aligned}
     \bm{emb}_{m\_e}&=\mathtt{MLP}(\mathtt{Concat}(\bm{emb}_{m}, \bm{emb}_{e*}), \\
     \bm{emb}_{p}&=\mathtt{Concat}(\mathtt{FFN}(\mathtt{SA(\bm{emb}_{m})}), \mathtt{FFN}(\mathtt{SA(\bm{emb}_{e*})}), \bm{emb}_{m\_e}), 
\end{aligned}
\end{equation}
where $\bm{emb}_{p}$ refer to portrait embeddings.
The latents $\bm{z}_{i}$ then perform cross-attention $CA(\cdot)$ with $\bm{emb}_{img}$ and $\bm{emb}_{p}$, respectively:
\begin{equation}\small
\label{eq:cross_attn_latent}
\begin{aligned}
     \bm{z}_{i}^{img}&=\mathtt{CA}(\bm{z}_{i}, \bm{emb}_{img}), \\
     \bm{z}_{i}^{p}&=\mathtt{CA}(\bm{z}_{i}, \bm{emb}_{p}), 
\end{aligned}
\end{equation}
To reduce the distance between distribution centers of $\bm{z}_{i}^{img}$ and $\bm{z}_{i}^{p}$, we ensure $\frac{\bm{z}_{i}^{img}-\bm{\mu}_{img}}{\bm{\sigma}_{img}}=\frac{\bm{z}_{i}^{p}-\bm{\mu}_{p}}{\bm{\sigma}_{p}}$, where where $\bm{\mu}_{img/p}$ and $\bm{\sigma}_{img/p}$ refer to the mean and standard deviation of $\bm{z}^{img/p}_{i}$, respectively. If the above equation holds, the distribution centers of these two features are nearly identical, thereby significantly enhancing identity stability across frames. Thus, we further normalize $\bm{z}_{i}^{p}$ and element-wise add it to $\bm{z}_{i}^{img}$ for facilitating identity consistency:
\begin{equation}\small
\label{eq:normalization}
\begin{aligned}
     \bar{\bm{z}}^{p}_{i}&=\frac{\bm{z}^{p}_{i}-\bm{\mu}_{p}}{\bm{\sigma}_{p}}\times\bm{\sigma}_{img}+\bm{\mu}_{img}, \\
     \bar{\bm{z}_{i}}&=\bar{\bm{z}}^{p}_{i}+\bm{z}^{img}_{i}.
\end{aligned}
\end{equation}

\subsection{Weighted Sliding-Window Strategy}
\label{sec: sling_window}

To improve the smoothness of video clip transition in long-length video generation, we propose a weighted sliding-window strategy during inference, as illustrated in Fig.\ref{fig:framework}(b).
Compared with conventional sliding window schemes~\cite{cheng2025wan_animate, xu2025hunyuanportrait, ji2025sonic}, we assign relative frame index-aware weights $\bm{W}=\{ \bm{w}_{i}=\frac{i}{v}| i=0,1,2,..,v\}$ on overlapping areas between adjacent windows and fuse overlapping latents (overlapping length $v$ between windows) via weighted summation:
\begin{equation}\small
\label{eq:sliding_window}
\begin{aligned}
    \bm{z}^{overlapp}_{i} &= \bm{W}*\bm{C}_{i} + (1-\bm{W})*\bm{C}_{i-1}
\end{aligned}
\end{equation}
where $\bm{z}^{overlapp}_{i}$ and $\bm{C}_{i}$ refer to overlapping latents and overlapping areas at the $i$-th window. Leveraging an arithmetic weighting function based on relative frame indices introduces a smooth blending effect in the transitions between adjacent windows. More details are described in Algorithm \ref{alg:sliding_window}.
$L$ and $l$ refer to the VAE-compressed total video length and window length. We set $v=5$ in our experiment.

\begin{algorithm}[t!]
\caption{Weighted Sliding-Window Strategy}
\label{alg:sliding_window}
\begin{algorithmic}[1]
\small 
\State \textbf{Input:} \text{DiT model }$\bm{\varepsilon}(\cdot)$, $\bm{z}_{T}^{[0,L]}$,{$\bm{emb}_{p}^{[0,L]}$, $T$, $l~(l<L)$, $v$} 
    \State \textbf{for} $t$ \text{in} $\mathtt{range}(T, 0, -1):$ \hfill $\triangleright$ $T$: denoising steps
        \State \hspace{0.5em} $\text{starting index }s=0$, $\text{ending index }e=s+l$
        \State \hspace{0.5em} $\text{previous ending index }e_{prev}=e$ \hfill $\triangleright$ $\bm{z}_{T}^{[0,L]}$: noised latents
        \State \hspace{0.5em} \textbf{while} $e\le L$ \textbf{do} \hfill
        \State \hspace{1.2em} $\bm{z}_{t-1}^{[s, e]}=\bm{\varepsilon}(\bm{z}_{t}^{[s, e]},\bm{emb}_{p}^{[s, e]},t)$
        \State \hspace{1.2em} \textbf{if} $s\ne0$ and $t \ne T$: 
        \State \hspace{1.8em} $\bm{w}=\mathtt{np.linspace}(0,1,\text{num\_samples=}v)$
        \State \hspace{1.8em} $\bm{z}_{t-1}^{[s,s+v]}=\bm{w}*\bm{z}_{t-1}^{[s,s+v]}+(1-\bm{w})*\bm{z}_{t-1}^{[e_{prev}-v,e_{prev}]}$
        \State \hspace{1.2em}  \textbf{if} $e<L$: \hfill $\triangleright$ \text{It covers the last clip case}
        \State \hspace{1.8em} $e_{prev}=e, s=s+(l-v), e=\mathtt{min}(s+l, L)$
        \State \hspace{1.2em}  \textbf{else}: \textbf{break}
    \State \textbf{return} $\bm{z}_{0}^{[0,L]}$
\end{algorithmic}
\vspace{-0.1cm}
\end{algorithm}

\subsection{Adaptive Latent Prediction Acceleration}
\label{sec: acceleration}
To accelerate denoising within each context window, we propose an Adaptive Latent Prediction Acceleration Mechanism that adaptively predicts future latents from historical latent differences, guided by the latent variation rate and inter-layer derivative magnitude ratio. 

Concretely, we first utilize a Taylor expansion to predict the latents at future timesteps, which can be formulated as
\begin{equation}\small
\label{eq:taylor_expansion}
\begin{aligned}
    \bm{f}(t)=\sum_{i=0}^{n}\frac{\bm{f}^{(i)}(a)}{i!}(t-a)^{i}+\bm{R}_{n+1}
\end{aligned}
\end{equation}
where $\bm{R}_{n+1}=\frac{\bm{f}^{(n+1)}(\xi )}{(n+1)!}(t-a)^{n+1}, \xi\in [t, a]$.
In our setting, $\bm{f}(\cdot)$ refers to a denoising DiT, and $\bm{a}$ is set to $t+k$, where $K$ is a timestep interval (set to 5) and $k \in \{1, ..., K-1\}$. Thus, we further derive the prediction as:
\begin{equation}\small
\label{eq:taylor_in_our_setting}
\begin{aligned}
    \bm{f}(t)=\sum_{i=0}^{n}\frac{\bm{f}^{(i)}(t+k)}{i!}(-k)^{i}+\bm{R}_{n+1}.
\end{aligned}
\end{equation}
To mitigate the extra computation and inference delay caused by differentiation, we employ finite differences to approximate derivatives, such as:
\begin{equation}\small
\label{eq:difference}
\begin{aligned}
    \bigtriangleup \bm{f}(t)&=\bm{f}(t+K)-\bm{f}(t), \\
    \bigtriangleup^{2} \bm{f}(t)&=\bigtriangleup\bm{f}(t+K)-\bigtriangleup\bm{f}(t).
\end{aligned}
\end{equation}
We then demonstrate the relationship $\bigtriangleup^{i}\bm{f}(t) \approx K^{i}\bm{f}^{(i)}(t)$ using mathematical induction.
In particular, we first explore their relationship for $i=1$:
\begin{equation}\small
\label{eq:stage-1}
\begin{aligned}
    \bigtriangleup \bm{f}(t)&=\bm{f}(t+K)-\bm{f}(t), \\
    &=[\bm{f}(t)+K\bm{f}^{'}(t)+\frac{K^{2}}{2}\bm{f}^{''}(t)+...]-\bm{f}(t) \\
    &=K\bm{f}^{'}(t)+\mathcal{O}(K^{2}) \approx K\bm{f}^{'}(t)
\end{aligned}
\end{equation}
We further assume that the above formulation holds for the $(i-1)$-th order difference ($\bigtriangleup^{i-1}\bm{f}(t)\approx \bm{K}^{i-1}\bm{f}^{(i-1)}(t)$). Thus, for the $i$-th order difference, we have $\bigtriangleup^{i}\bm{f}(t)=\bigtriangleup(\bigtriangleup^{i-1}\bm{f}(t))=\bigtriangleup^{i-1}\bm{f}(t+K)-\bigtriangleup^{i-1}\bm{f}(t)$. we further simplify $\bigtriangleup^{i-1}\bm{f}(t+K)$ as:
\begin{equation}\small
\label{eq:simplification}
\begin{aligned}
    \bigtriangleup^{i-1}\bm{f}(t+K)&=\bigtriangleup^{i-1}\bm{f}(t)+K\cdot\frac{d\bigtriangleup^{i-1}\bm{f}(t)}{dt}+\mathcal{O}(K^{2}), \\
    &\approx K^{i-1}\bm{f}^{(i-1)}(t)+K\cdot K^{i-1}\cdot\bm{f}^{(i)}(t) \\
    &=K^{i-1}\bm{f}^{(i-1)}(t)+K^{i}\cdot\bm{f}^{(i)}(t)
\end{aligned}
\end{equation}
Thus, we substitute Eq. \ref{eq:simplification} into $\bigtriangleup^{i}\bm{f}(t)$ to obtain:
\begin{equation}\small
\label{eq:substitute_final}
\begin{aligned}
    \bigtriangleup^{i}\bm{f}(t)&\approx K^{i-1}\bm{f}^{(i-1)}(t)+K^{i}\cdot\bm{f}^{(i)}(t)-K^{i-1}\bm{f}^{(i-1)}(t), \\
    &=K^{i}\bm{f}^{(i)}(t)
\end{aligned}
\end{equation}
Therefore, $\bigtriangleup^{i}\bm{f}(t)\approx K^{i}\bm{f}^{(i)}(t)$ has been fully verified, and we substitute it into Eq. \ref{eq:taylor_in_our_setting} to obtain the converted prediction:
\begin{equation}\small
\label{eq:converted_prediction}
\begin{aligned}
    \bm{f}(t)=\sum_{i=0}^{n}\frac{\bigtriangleup^{i}\bm{f}(t+k)}{i!K^{i}}(-k)^{i}+\bm{R}_{n+1}.
\end{aligned}
\end{equation}
The DiT only needs to fully denoise the latents at $\{ t+(n+1)K,...,t+2K, t+K\}$ timesteps.

However, as we observe that portrait animation contains intricate and large-amplitude facial motion patterns, it results in dramatic fluctuations in latent distribution across different timesteps. Thus, fixed-formatting prediction methods suffer from ID inconsistency due to their inaccurate predicted latents. To address this issue, we design two dynamic functions that adaptively refine predictions, ensuring ID stability and efficient acceleration. In particular, we first calculate the latent variation at specific timestep ($\mathtt{\sigma}(t)=\frac{d\bm{f}(t)}{dt}$) and the average latent variation across timesteps ($\mathtt{\sigma}_{avg}(T^{'})=\frac{1}{T-T^{'}}\int_{T^{'}}^{T}\mathtt{\sigma}(t)dt$). $T$ is the total timestep number. We then define the first dynamic function based on the latent variation rate at each timestep as follows:
\begin{equation}\small
\label{eq:first_dynamic_function}
\begin{aligned}
    \bm{s}(t)=(\frac{\mathtt{\sigma}(t)}{\mathtt{\sigma}_{avg}(t)})^{\alpha},
\end{aligned}
\end{equation}
where $\alpha \in [0.5, 1.5]$ (set to 1.5 in our experiments).
At early timesteps, the latents vary rapidly ($\bigtriangleup^{i}\bm{f}(t) \uparrow$), necessitating a larger $K$ to compensate for the pronounced variations. As the diffusion process enters later timesteps, latent updates become more gradual ($\bigtriangleup^{i}\bm{f}(t) \downarrow$), so $K$ is reduced to prevent excessive amplification of $\bigtriangleup^{i}\bm{f}(t)$.
We then define the second function based on the derivative magnitude among diffusion layers:
\begin{equation}\small
\label{eq:second_dynamic_function}
\begin{aligned}
    \bm{r}(t, l, i)&=\frac{\mathtt{E}[||\bm{f}^{(i)}(t, l)||]}{\mathtt{E}[||\bm{f}^{(i)}(t, avg)||]}, \\
    \bm{w}(t, l, i)&=\frac{1}{\sqrt{\bm{r}(t, l, i)}},
\end{aligned}
\end{equation}
where $l$, $\bm{f}^{(i)}(t, l)$, and $\mathtt{E}[||\bm{f}^{(i)}(t, avg)||]$ refer to the DiT layer index, the $i$-th order derivative at the $l$-th DiT layer, and the average derivative magnitude across all layers. 
By dynamically adjusting the mapping between finite differences and high-order derivatives across layers, $\bm{w}(\cdot)$ addresses prediction errors caused by large derivative-magnitude gaps across layers at the same timestep. 
For lower layers that capture texture and edges and are more noise-sensitive ($\bm{f}^{(i)}(t, l) \uparrow$), high-order derivatives fluctuate sharply, causing the finite-difference approximation to be underestimated. Thus, when $\bm{r}(\cdot)>1$, the scaling factor is reduced to avoid excessive amplification. 
For higher layers that model stable global structures ($\bm{f}^{(i)}(t, l) \downarrow$), derivatives vary smoothly and the approximation is overestimated. When $\bm{r}(\cdot)<1$, the scaling factor is increased to counteract the insufficient finite-difference magnitude.

These two dynamic functions can significantly tackle the fluctuations in latent distribution across different timesteps due to complex facial motion patterns.
We further refine $\bigtriangleup^{i}\bm{f}(t)\approx K^{i}\bm{f}^{(i)}(t)$ using Eq. \ref{eq:first_dynamic_function} and Eq. \ref{eq:second_dynamic_function}:
\begin{equation}\small
\label{eq:final_convertion}
\begin{aligned}
    \bigtriangleup^{i}\bm{f}(t, l)\approx K^{i}\cdot \bm{w}(t, l, i) \cdot \bm{s}(t) \cdot \bm{f}^{(i)}(t, l),
\end{aligned}
\end{equation}
Therefore, we substitute Eq. \ref{eq:final_convertion} into Eq. \ref{eq:converted_prediction} to obtain the final prediction formulation as follows:
\begin{equation}\small
\label{eq:final_predicition}
\begin{aligned}
    \bm{f}(t, l)=\bm{f}(t+k, l)+\sum_{i=1}^{n}\frac{\bigtriangleup^{i}\bm{f}(t+k, l)\cdot(-k)^{i}}{i!\cdot K^{i}\cdot\bm{w}(t+k, l, i)\cdot\bm{s}(t+k)},
\end{aligned}
\end{equation}
Notably, we omit $\bm{R}_{n+1}$ for brevity.

\subsection{Training}
\label{sec: training}

FlashPortrait is trained using reconstruction loss, with only the attention modules of the DiT being trainable. To improve face-region fidelity, we apply face and lip masks $\bm{M}_{face}$ and $\bm{M}_{lip}$, which are obtained from the input video frames via MediaPipe~\cite{lugaresi2019mediapipe}:
\begin{equation}\small
\label{eq:loss}
\begin{aligned}
     \mathcal{L}=\mathbb{E}_{\theta}(\left \| (\bm{z}_{gt}-\bm{z}_{\varepsilon})\odot  (1+\bm{M}_{face}+\bm{M}_{lip})  \right \|^{2})
\end{aligned}
\end{equation}
where $\bm{z}_{gt}$ and $\bm{z}_{\varepsilon}$ refer to diffusion latents and denoised latents. 
This loss facilitates more targeted and effective learning on different facial components.
\section{Experiments}
\label{sec:experiments}

\begin{table*}[t!]\small
\caption{Quantitative comparisons on Voxceleb2\&Vfhq and Hard100. In the table elements $a$ / $b$, $a$ and $b$ refer to the result on the Voxceleb2\&Vfhq and Hard100, respectively. The average video duration of Voxceleb2\&Vfhq is 10 seconds, while Hard100 is 1 minute.
LMD/APD multiplied by $10^{-3}$ and AED multiplied by $10^{-2}$. Speed refers to the inference latency for a 20-second 480×832 video.
}
\vspace{-0.25in}
\begin{center}
\renewcommand\arraystretch{1.1}
\scalebox{0.85}{
\begin{tabular}{l|ccccccccc}
\toprule
Model           & FID$\downarrow$                  & FVD$\downarrow$                    & PSNR$\uparrow$        & SSIM$\uparrow$      & LMD$\downarrow$                & AED$\downarrow$                  & APD$\downarrow$                  & MAE$\downarrow$                 & Speed$\downarrow$ \\ \midrule
LivePortrait~\cite{guo2024liveportrait}    & 83.21/143.72         & 492.46/584.13          & 31.53/20.48 & 0.74/0.72 & 8.75/10.02         & 28.64/50.43          & 27.57/43.68          & 10.22/20.17         & \textbf{108s}   \\
Skyreels-A1~\cite{qiu2025skyreels}     & 70.48/168.24         & 386.24/758.25          & 31.04/15.42 & 0.76/0.68 & 5.34/10.45         & 21.26/46.11          & 20.43/48.35          & 9.17/21.23          & 504s  \\
FollowYE~\cite{ma2024follow}        & 76.45/221.36         & 425.18/845.69          & 30.65/21.26 & 0.69/0.73 & 10.28/9.47         & 23.69/58.62          & 21.32/38.12          & 14.81/23.85         & 648s  \\
X-Portrait~\cite{xie2024x_portrait}      & 85.13/164.68         & 413.53/814.46          & 30.82/18.38 & 0.73/0.74 & 9.47/8.82          & 22.81/52.94          & 20.98/39.53          & 9.49/21.04          & 1636s \\
HunyuanPortrait~\cite{xu2025hunyuanportrait} & 73.62/157.93         & 366.72/882.54          & 31.93/16.63 & 0.78/0.69 & 6.02/8.64          & 20.75/49.95          & 20.14/41.08          & 8.85/20.48          & 1602s \\
FantasyPortrait~\cite{wang2025fantasyportrait} & 65.27/165.71         & 328.93/723.57          & 32.48/16.47 & 0.80/0.71 & 5.24/8.49          & 19.66/45.34          & 19.36/36.67          & 7.64/19.87          & 4339s \\
Wan-Animate~\cite{cheng2025wan_animate}     & 65.20/143.61         & 336.12/695.48          & \textbf{32.54}/18.13 & \textbf{0.82}/0.72 & 5.17/7.98          & 19.54/42.98          & 19.15/35.06          & 7.88/20.08          & 2298s \\ \midrule
Ours            & \textbf{65.18/62.33} & \textbf{320.47/340.21} & 32.36/\textbf{26.16} & 0.79/\textbf{0.82} & \textbf{4.90/5.26} & \textbf{15.19/29.68} & \textbf{14.46/24.40} & \textbf{5.93/12.54} & 720s  \\ \bottomrule
\end{tabular}
}
\end{center}
\vspace{-0.3in}
\label{table:quantitative_comparisons}
\end{table*}

\subsection{Implementation Details}
Our training dataset is comprised of three components: Hallo3~\cite{cui2025hallo3}, Celebv-HQ~\cite{zhu2022celebvhq}, and a collection of internet-sourced videos, amounting to a total of 2000 hours. We first follow prior work~\cite{xu2025hunyuanportrait} and test FlashPortrait on the Voxceleb2~\cite{chung2018voxceleb2} and Vfhq~\cite{xie2022vfhq} datasets. As previous works do not open-source their testing datasets, we randomly select 100 videos (5-20 seconds long) from both Voxceleb2 and Vfhq. In addition, we conduct robustness assessments by testing our model on 100 unseen videos (1-3 minutes in length, FPS=30) from the internet, labeled as the Hard100. Our DiT model leverages pre-trained weights from Wan2.1-I2V-14B~\cite{wan2025}. The training process spans 20 epochs, utilizing 200 NVIDIA H100 80GB GPUs with a batch size of 1 per GPU. We set learning rate$=1e-5$, $K=5$, and $n=3$.

\subsection{Comparison with State-of-the-Art Methods}
\textbf{Quantitative results.}
Following~\cite{wang2025fantasyportrait, xu2025hunyuanportrait}, in the self-reenactment, we utilize FID~\cite{heusel2017gans}, FVD~\cite{unterthiner2018towards}, PSNR, and SSIM~\cite{wang2004image} to assess the quality of synthesized images and videos. We leverage LMD~\cite{lugaresi2019mediapipe} and MAE~\cite{han2024face} to evaluate the expression motion accuracy and eye movement accuracy. In the cross-reenactment, we use AED~\cite{siarohin2019first} and APD~\cite{siarohin2019first} to assess the accuracy of expression and head movement.
We compare with recent portrait animation models, including GAN-based models (LivePortrait~\cite{guo2024liveportrait}) and diffusion-based models (UNet-based: FollowYE~\cite{ma2024follow}, X-Portrait~\cite{xie2024x_portrait}; DiT-based: Skyreels-A1~\cite{qiu2025skyreels}, HunyuanPortrait~\cite{xu2025hunyuanportrait}, FantasyPortrait~\cite{wang2025fantasyportrait}, Wan-Animater~\cite{cheng2025wan_animate}).
We perform quantitative comparisons with the above competitors on Voxceleb2~\cite{chung2018voxceleb2} $\&$ Vfhq~\cite{xie2022vfhq} and Hard100, as shown in Table \ref{table:quantitative_comparisons}.
We observe that although all competitors experience a notable performance drop in long video generation, FlashPortrait still outperforms them in expression/eye-motion accuracy, video fidelity, and single-frame quality, while maintaining relatively high quality for short video generation.
Furthermore, FlashPortrait achieves the fastest inference speed among all DiT-based methods~\cite{qiu2025skyreels, xu2025hunyuanportrait, wang2025fantasyportrait, cheng2025wan_animate} and delivers the best performance on both test sets.
Specifically, Wan2.1-14B-based FlashPortrait surpasses the best competitor Wan2.2-14B-based Wan-Animate by 30.9\%/30.4\%/37.5\% in AED/APD/MAE on Hard100, while achieving 3$\times$ faster inference speed.

\begin{figure*}[t!]
\begin{center}
\includegraphics[width=0.98\linewidth]{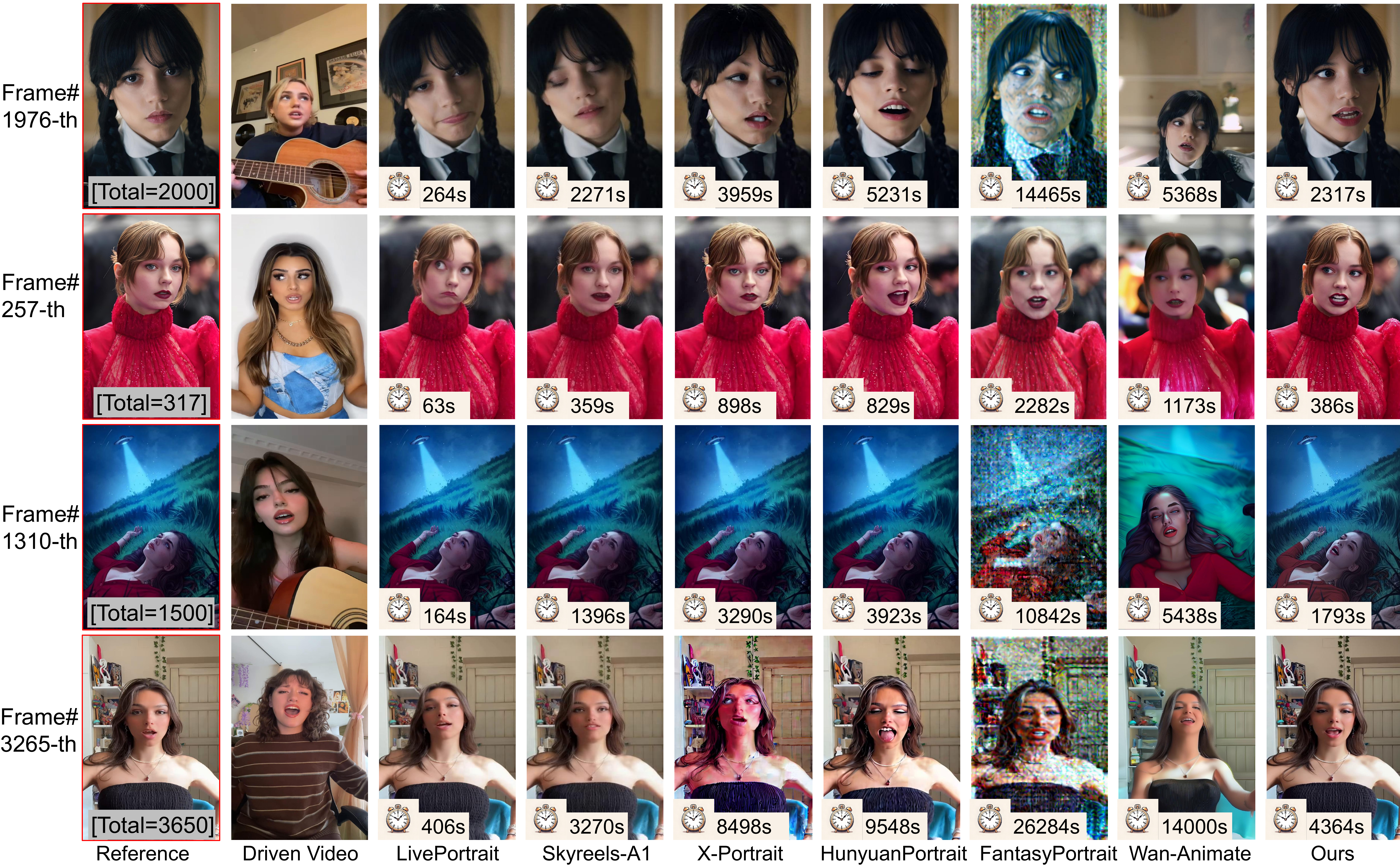}
\end{center}
\vspace{-0.65cm}
   \caption{Qualitative comparisons with state-of-the-art methods. [Total=X] refers to the total frame number of the video.}
\label{fig:comparison}
\vspace{-0.5cm}
\end{figure*}

\begin{figure}[t!]
\begin{center}
\includegraphics[width=1\linewidth]{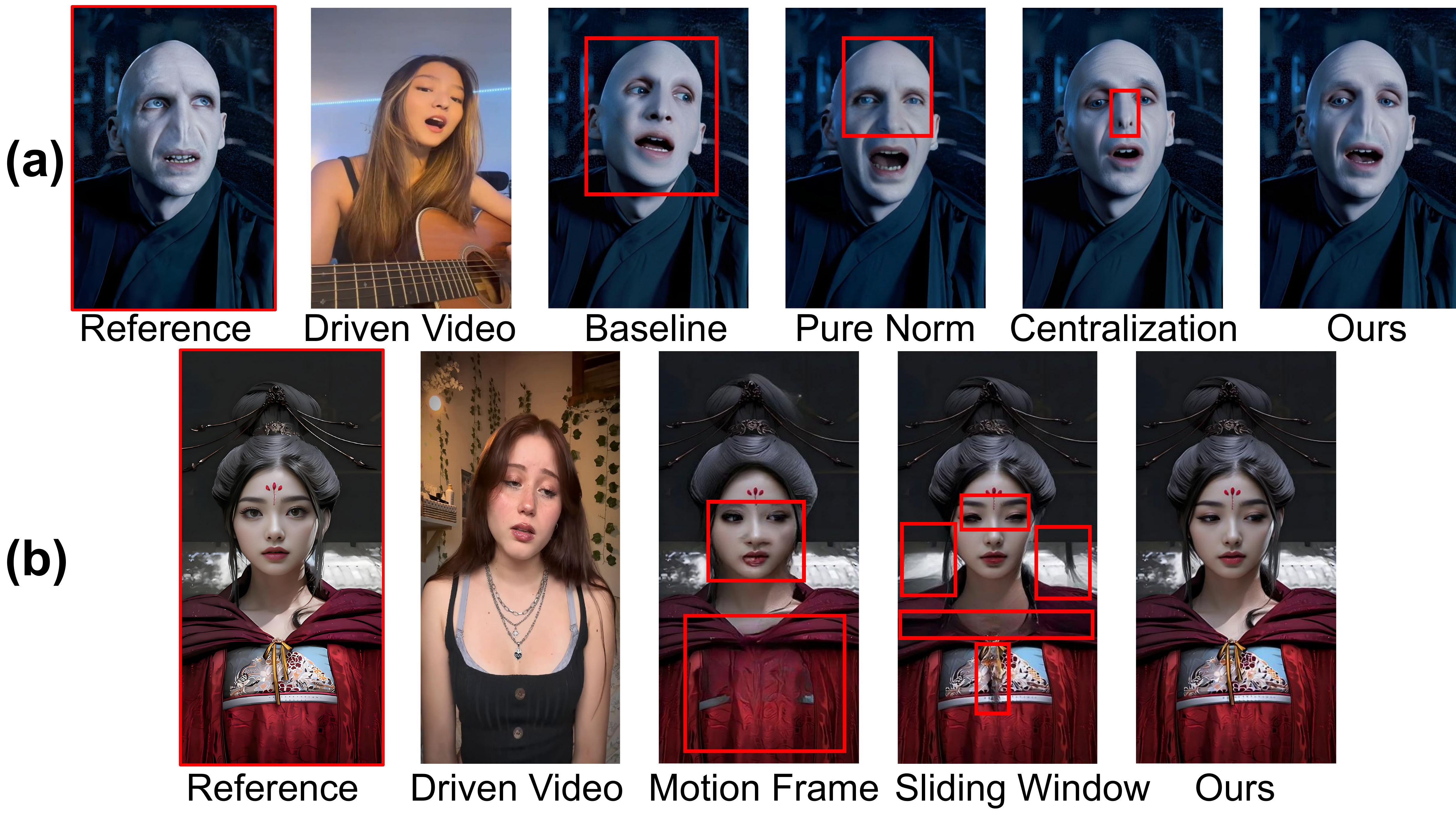}
\end{center}
\vspace{-0.6cm}
   \caption{Ablation study on normalization and long video.
   }
\label{fig:ablation_norm_window}
\vspace{-0.35cm}
\end{figure}

\noindent\textbf{Qualitative Results.}
The qualitative results are shown in Fig. \ref{fig:comparison}.
Notably, each driven video is filled with intricate expression patterns, while the references include intricate details of appearances. 
As FantasyPortrait~\cite{wang2025fantasyportrait} doesn't contain long video generation mechanisms, we adopt the sliding-window strategy from Wan-Animate~\cite{cheng2025wan_animate} to it.
In particular, LivePortrait~\cite{guo2024liveportrait} and Skyreels-A1~\cite{qiu2025skyreels} preserve identity, yet struggle to faithfully transfer reference facial expressions from the driving video, particularly in capturing eye dynamics and mouth movements.
X-Portrait~\cite{xie2024x_portrait} suffers from significant face distortion.
As Wan-Animate relies on explicit poses, body misalignment between the reference and driven videos greatly degrades performance, especially when its default alignment fails.
HunyuanPortrait~\cite{xu2025hunyuanportrait}, FantasyPortrait~\cite{wang2025fantasyportrait}, and Wan-Animate~\cite{cheng2025wan_animate} exhibit severe color drift, face/body distortion, and stochastic facial expression changes that deviate from the driving video guidance once the video length surpasses 30 seconds, along with substantial inference latency.
In contrast, our FlashPortrait accurately animates images based on the given video while preserving reference identities even after generating 3000+ frames, highlighting the superiority of our model in generating vivid, infinite-length videos. It further achieves the most favorable trade-off between inference speed and visual quality among all DiT-based competitors~\cite{qiu2025skyreels, xu2025hunyuanportrait, wang2025fantasyportrait, cheng2025wan_animate}.

\subsection{Ablation Study}

\begin{table}[t!]\small
\caption{Ablation study on Normalized Facial Expression Blocks. $\bar{\bm{z}_{i}}$ in the Baseline, Pure Norm, and Centralization refer to $\bm{z}^{p}_{i}+\bm{z}_{i}^{img}$, $\frac{\bm{z}^{p}_{i}-\bm{\mu}_{p}}{\bm{\sigma}_{p}}+\bm{z}_{i}^{img}$, and $\frac{\bm{z}^{p}_{i}-\bm{\mu}_{p}}{\bm{\sigma}_{p}}+\frac{\bm{z}^{img}_{i}-\bm{\mu}_{img}}{\bm{\sigma}_{img}}$, respectively.
}
\vspace{-0.25in}
\begin{center}
\renewcommand\arraystretch{1.1}
\scalebox{0.85}{
\begin{tabular}{l|ccc}
\toprule
Model          & AED$\downarrow$            & APD$\downarrow$            & MAE$\downarrow$            \\ \midrule
Baseline       & 44.78          & 36.87          & 19.73          \\
Pure Norm      & 38.42          & 32.64          & 17.25          \\
Centralization & 33.76          & 27.31          & 14.66          \\ \midrule
Ours           & \textbf{29.68} & \textbf{24.40} & \textbf{12.54} \\ \bottomrule
\end{tabular}
}
\end{center}
\label{table:ablation_norm}
\vspace{-0.25in}
\end{table}

\textbf{Normalization.}
We conduct an ablation study to demonstrate the contributions of Normalized Facial Expression Blocks in FlashPortrait, as shown in Table \ref{table:ablation_norm} and Fig. \ref{fig:ablation_norm_window}(a).
Notably, all quantitative ablation studies are on the Hard100 dataset. 
We can see that the Pure Norm and Centralization fail to fully maintain the facial details and identity consistency, as they do not fundamentally narrow the distribution gap between the latent space and facial embeddings.
By contrast, our Normalized Facial Expression Blocks can ensure high facial quality by integrating the mean and the standard deviation from both cross-attention features, significantly reducing the distance between distribution centers of latents and raw facial embeddings.

\begin{table}[t!]\small
\caption{Ablation study on long portrait animation methods.
}
\vspace{-0.25in}
\begin{center}
\renewcommand\arraystretch{1.1}
\scalebox{0.85}{
\begin{tabular}{l|ccc}
\toprule
Model          & AED$\downarrow$            & APD$\downarrow$            & MAE$\downarrow$            \\ \midrule
Motion Frame~\cite{cui2025hallo3}   & 37.25          & 30.71          & 17.67          \\
Sliding Window~\cite{cheng2025wan_animate} & 36.44          & 28.12          & 14.90          \\ \midrule
Ours           & \textbf{29.68} & \textbf{24.40} & \textbf{12.54} \\ \bottomrule
\end{tabular}
}
\end{center}
\label{table:ablation_long}
\vspace{-0.25in}
\end{table}

\noindent\textbf{Long Video.} We conduct an ablation study on expression/eye motion accuracy in the long portrait animation, as shown in Fig. \ref{fig:ablation_norm_window}(b) and Table \ref{table:ablation_long}. Compared with the motion frame~\cite{cui2025hallo3} and conventional sliding window~\cite{cheng2025wan_animate}, our Weighted Sliding-Window Strategy dynamically fuses adjacent context windows through weighted aggregation in their overlapping regions, enabling seamless transitions between sub-clips, thereby significantly improving the synthesized long animation quality.

\begin{figure}[t!]
\begin{center}
\includegraphics[width=1\linewidth]{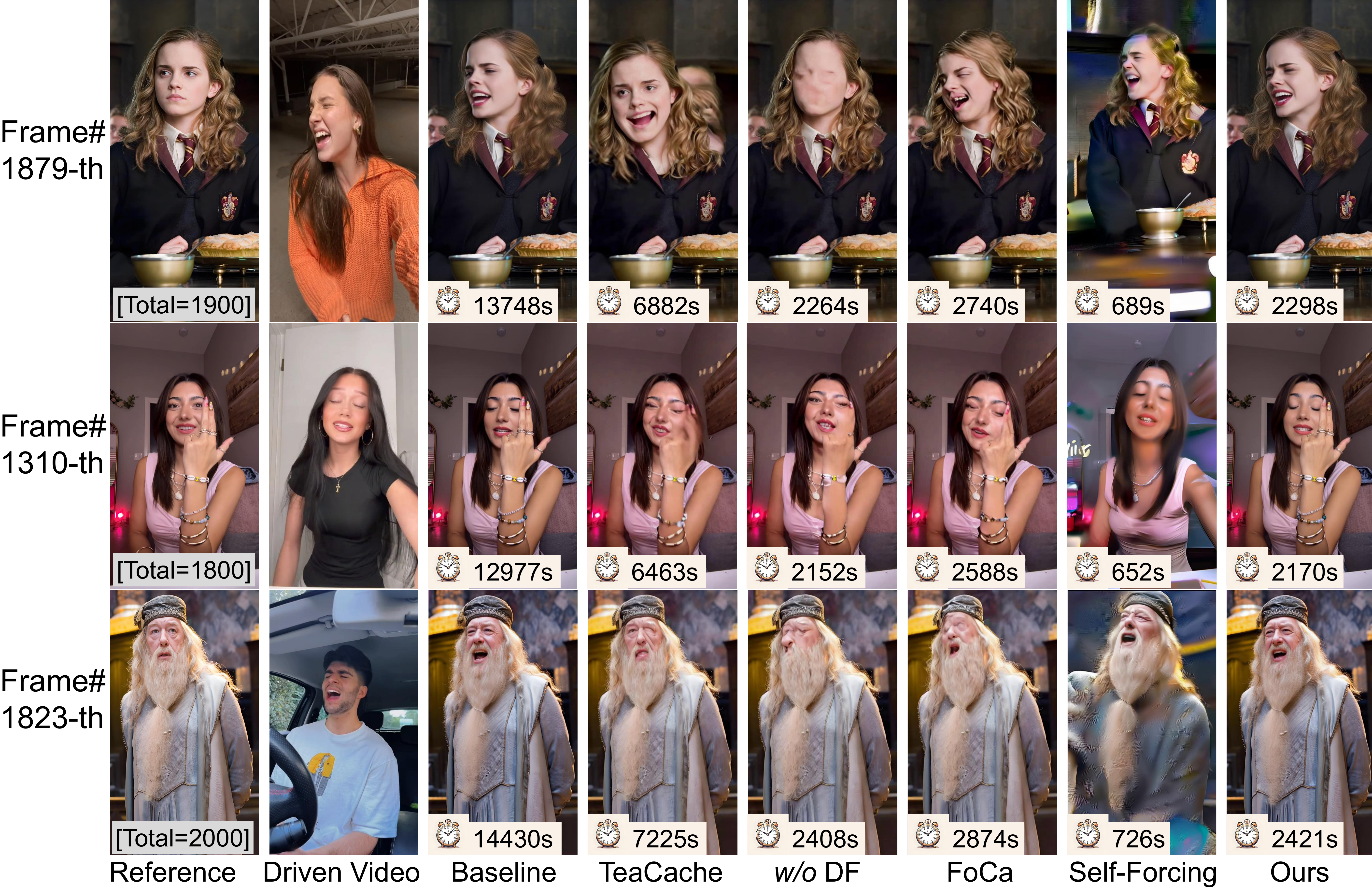}
\end{center}
\vspace{-0.6cm}
   \caption{Ablation study on acceleration.
   }
\label{fig:ablation_acceleration}
\vspace{-0.45cm}
\end{figure}

\begin{table}[t!]\small
\caption{Ablation study on acceleration methods. The baseline and \textit{w/o} Dynamic Functions remove our acceleration strategy in FlashPortrait and $\bm{w}(l, i) \cdot \bm{s}(\bm{t})$ during latent prediction. Speed refers to the inference latency for a 20-second 480×832 video.
}
\vspace{-0.25in}
\begin{center}
\renewcommand\arraystretch{1.1}
\scalebox{0.85}{
\begin{tabular}{l|cccc}
\toprule
Model                 & AED$\downarrow$            & APD$\downarrow$            & MAE$\downarrow$            & Speed$\downarrow$         \\ \midrule
Baseline              & 29.12          & 23.86          & 12.37          & 4328s         \\
TeaCache~\cite{liu2025teacache}              & 33.94          & 27.62          & 15.06          & 2164s         \\
\textit{w/o} Dynamic Functions & 42.66          & 35.98          & 19.63          & 682s          \\
FoCa~\cite{zheng2025forecast}                  & 37.47          & 32.96          & 17.88          & 862s          \\
Self-Forcing~\cite{huang2025selfforcing}          & 52.85          & 39.32          & 20.79          & \textbf{266s} \\ \midrule
Ours                  & \textbf{29.68} & \textbf{24.40} & \textbf{12.54} & 720s          \\ \bottomrule
\end{tabular}
}
\end{center}
\label{table:ablation_acceleration}
\vspace{-0.4in}
\end{table}

\noindent\textbf{Acceleration.}
To validate the significance of our Adaptive Latent Prediction Acceleration, we conduct an ablation study on various types of acceleration methods~\cite{liu2025teacache, zheng2025forecast, huang2025selfforcing, liu2025reusing}, as shown in Table \ref{table:ablation_acceleration} and Fig. \ref{fig:ablation_acceleration}.
\textit{w/o} DF refers to \textit{w/o} Dynamic Functions.
We have the following observations: 
(1) While TeaCache~\cite{liu2025teacache} enables acceleration with minimal performance loss compared to the baseline, its speed-up is limited to at most 2$\times$.
(2) \textit{w/o} Dynamic Functions encounters significant performance deterioration. The plausible reason is that portrait animation involves complex, large-amplitude facial motions, which cause substantial fluctuations in the latent distribution across timesteps. Thus, fixed-pattern prediction approaches encounter identity inconsistency from inaccurate latent predictions.
(3) Although FoCa~\cite{zheng2025forecast} and Self-Forcing~\cite{huang2025selfforcing} achieve extremely high inference acceleration ratios (5$\times$–20$\times$), their synthesized videos often suffer from severe artifacts and identity inconsistency, especially when the driven video exhibits large-amplitude facial expressions and eye movements. The fundamental reason is that such dramatic expression dynamics in portrait animation make FoCa’s calibration unable to capture latent variations across timesteps accurately. Meanwhile, Self-Forcing relies on an autoregressive DMD-based~\cite{yin2024one, yin2024improved} 4-step student model, where the limited 4-step inference heavily overlooks complex expression motion modeling. Thus, Self-Forcing is generally suitable only for relatively static videos.
(4) Our acceleration strategy outperforms existing alternatives by achieving a 6$\times$ speed-up with negligible performance loss. Even under large expression and eye movements, it synthesizes high-quality videos without noticeable artifacts and preserves strong identity consistency.

We further conduct an ablation study on $\bm{K}$ and $\bm{n}$ in our Adaptive Latent Prediction Acceleration, as shown in Table \ref{table:ablation_acceleration_n_k}.
We can observe that a larger $\bm{K}$ yields higher acceleration but causes stronger degradation, especially when $\bm{K}>5$. Increasing $\bm{n}$ enhances latent prediction accuracy but reduces speed, and the benefit becomes marginal once $\bm{n}>3$. Consequently, $\bm{K}=5$ and $\bm{n}=3$ achieve the best quality-speed trade-off.
More ablation studies are depicted in the Sec.D of the Supp.

\begin{table}[t!]\small
\caption{Ablation study on $\bm{K}$ and $\bm{n}$ in our acceleration.
}
\vspace{-0.25in}
\begin{center}
\renewcommand\arraystretch{1.1}
\scalebox{0.85}{
\begin{tabular}{l|cccc}
\toprule
Model     & AED$\downarrow$   & APD$\downarrow$   & MAE$\downarrow$   & Speed$\downarrow$ \\ \midrule
n=1 (K=5) & 34.63 & 28.78 & 15.80 & 483s  \\
n=2 (K=5) & 32.25 & 26.54 & 13.97 & 569s  \\
n=4 (K=5) & 29.60 & 24.18 & 12.48 & 1025s \\ \midrule
K=2 (n=3) & 29.20 & 23.92 & 12.41 & 2116s \\
K=8 (n=3) & 44.21 & 36.68 & 19.47 & 295s  \\ \midrule
n=3, K=5  & 29.68 & 24.40 & 12.54 & 720s  \\ \bottomrule
\end{tabular}
}
\end{center}
\label{table:ablation_acceleration_n_k}
\vspace{-0.25in}
\end{table}

\subsection{Applications and User Study}
\noindent\textbf{Full Body Portrait Animations.}
We conduct a qualitative experiment on our FlashPortrait in full-body portrait animation. The results are shown in Sec. E of the Supp.
We can see that our FlashPortrait can handle full-body portrait animation in high-fidelity while preserving identities even in the presence of large-scale expression motions.

\begin{figure}[t!]
\begin{center}
\includegraphics[width=1\linewidth]{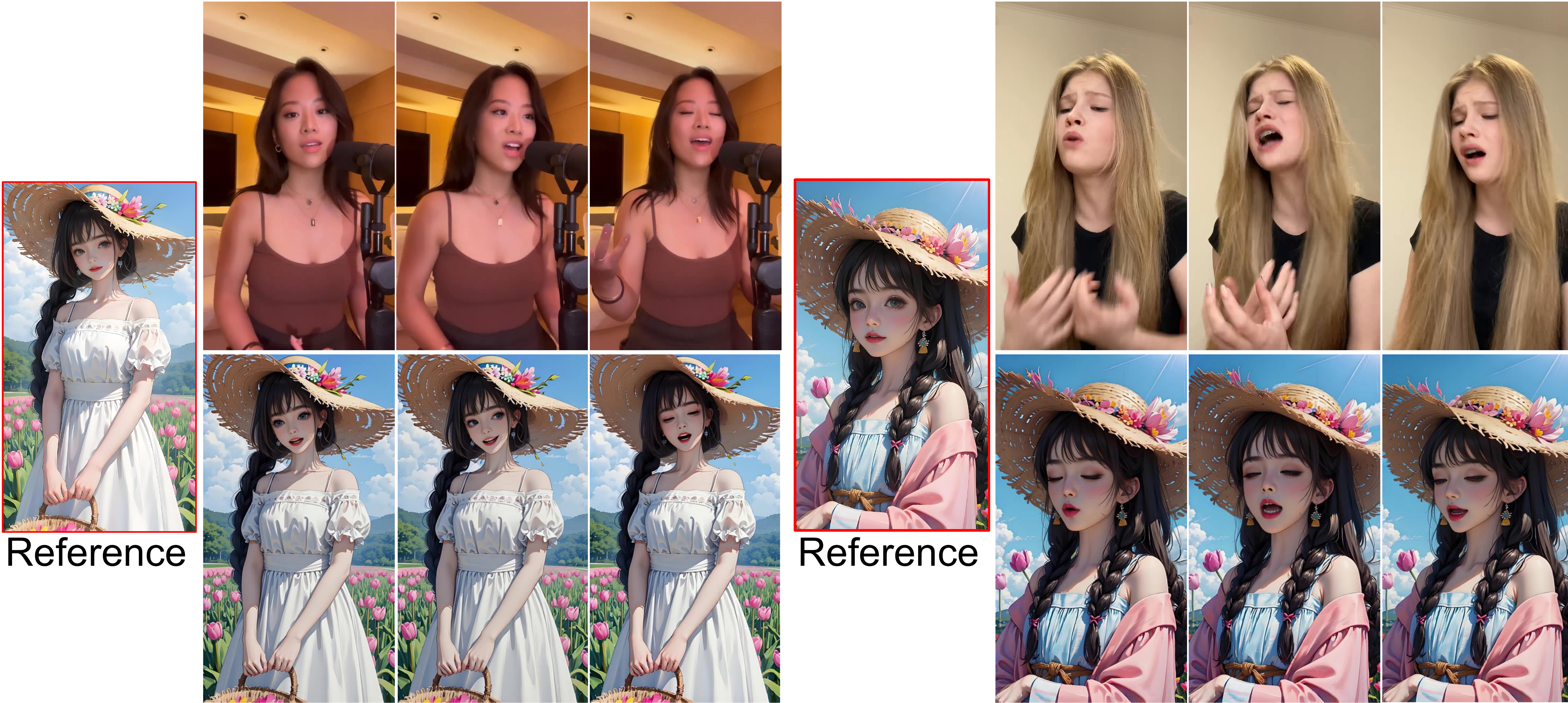}
\end{center}
\vspace{-0.6cm}
   \caption{Cartoon portrait animation results.
   }
\label{fig:cartoon}
\vspace{-0.35cm}
\end{figure}

\noindent\textbf{Cartoon Portraits.}
To validate the robustness of our FlashPortrait, we experiment on cartoon portrait animation, as shown in Fig. \ref{fig:cartoon}. We can observe that our model can synthesize natural cartoon portrait animation with rich facial expressions based on the driven video.

\noindent\textbf{Video Length.} To validate the performance of FlashPortrait in long portrait animation, we experiment on an extremely long case (3 minutes+, FPS=30), as shown in Sec. F of the Supp. It indicates that the video fidelity, expression/eye motion synchronization, and identity consistency remain stable without significant degradation, even after generating 5400+ frames.
Theoretically, FlashPortrait is capable of generating hours of video at high speed while maintaining stable quality without significant degradation.

\noindent\textbf{User Study.} To assess perceptual quality in a subjective way, we conducted a user study involving 30 curated samples. The participants, primarily university students and faculty, are first shown the reference image and its driven video. They then view two synthesized results—one from FlashPortrait and another from a competing method—presented in random order. Participants are then asked to answer the following questions: 
L-A/A-A/B-A/I-A: ``Which one has better facial expression motion/foreground appearance/background/identity alignment with the driven video/reference". The results in Table~\ref{table:user_study} demonstrate that FlashPortrait is consistently preferred in all evaluated aspects.

\begin{table}[t!]\small
\caption{User preference of FlashPortrait compared with other competitors. Higher indicates users prefer more to our model.
}
\vspace{-0.25in}
\begin{center}
\renewcommand\arraystretch{1.1}
\scalebox{0.85}{
\begin{tabular}{l|cccc}
\toprule
Model           & L-A    & A-A    & B-A    & I-A    \\ \midrule
LivePortrait~\cite{guo2024liveportrait}    & 95.4\% & 97.2\% & 98.5\% & 97.9\% \\
HunyuanPortrait~\cite{xu2025hunyuanportrait} & 94.8\% & 96.4\% & 98.2\% & 97.6\% \\
FantasyPortrait~\cite{wang2025fantasyportrait} & 95.2\% & 95.8\% & 97.7\% & 96.8\% \\
Wan-Animate~\cite{cheng2025wan_animate}     & 92.8\% & 93.7\% & 97.4\% & 96.5\% \\ \bottomrule
\end{tabular}
}
\end{center}
\label{table:user_study}
\vspace{-0.3in}
\end{table}
\section{Conclusion}
\label{sec:conclusion}
In this paper, we propose FlashPortrait, equipped with specialized training and inference mechanisms that enable infinite-length, ID-preserving portrait animation, while achieving up to a 6$\times$ acceleration in inference speed.
FlashPortrait first utilized an off-the-shelf model to obtain identity-agnostic facial expression features. To improve ID stability, FlashPortrait introduced a Normalized Facial Expression Block to refine expression features. 
In inference, to ensure the long video's smoothness and ID consistency, FlashPortrait proposed a Weighted Sliding-Window Strategy. In each context window, FlashPortrait further introduced an Adaptive Latent Prediction Acceleration Mechanism to skip several denoising steps, thereby achieving 6$\times$ speed acceleration.
Experimental results across various datasets demonstrated the superiority of our model in synthesizing infinite-length ID-preserving portrait animations with significantly faster speed.
{
    \small
    \bibliographystyle{ieeenat_fullname}
    \bibliography{main}
}
\clearpage
\setcounter{page}{1}
\appendix
\section{Supplementary Material}

\subsection{Evaluation Metrics}
Following previous portrait animation evaluation settings, we implement numerous quantitative evaluation metrics, including FID, FVD, LMD, AED, APD, and MAE, to compare our FlashPortrait with current state-of-the-art portrait animation models.
The details of the above metrics are described as follows:
\begin{enumerate}[label=(\arabic*)]
    \item FID  refers to measure the similarity in feature distribution between synthesized and real images, employing Inception v3 features.
    \item FVD refers to evaluate temporal coherence through features extracted from a pretrained model~\cite{unterthiner2019fvd}.
    \item LMD refers to measure the accuracy of synthesized facial expressions. The landmarks are extracted using Mediapipe. It computes the average Euclidean distance between the facial landmarks of the reference and synthesized images.
    \item AED refers to calculate the Manhattan distance of expression from SMIRK~\cite{retsinas20243d}, with lower
 values indicating better expression.
    \item APD calculate the Manhattan distance of  pose parameters from SMIRK~\cite{retsinas20243d}, with lower values indicating better pose similarity.
    \item MAE refers to measure the Mean Angular Error on the eye movement accuracy.
\end{enumerate}

\subsection{Preliminaries}

Diffusion models function through a stochastic process, consisting of two main phases: a forward diffusion step and a reverse denoising step for controlled noise addition and removal. In the forward process, Gaussian noise is gradually introduced to a data sample $\bm{x}_{0}\sim\bm{p}_{\text{data}}$, where $\bm{p}_{\text{data}}$ represents the underlying data distribution. This is done as follows, based on the Rectified Flow method~\cite{lipman2022flow}:
\begin{equation}\small
\label{eq:forward_diffusion}
\begin{aligned}
    \bm{x}_{t}=(1-t)\bm{x}_{0}+t\bm{x}_{1},
\end{aligned}
\end{equation}
where $t \in [0,1]$ denotes the timestep.
After $\bm{T}$ diffusion steps, the original data sample $\bm{x}_{0}$ is transformed into pure Gaussian noise $\bm{x}_{1}\sim\mathcal{N}(0,I)$.
In the reverse denoising process, the diffusion model $\bm{\varepsilon}_{\theta}(\bm{x}_{t},t)$ is trained to predict the velocity $(\bm{x}_{1}-\bm{x}_{0})$ conditioned on the noisy latents $x_{t}$ and the timestep $t$. To train the model, the Mean Squared Error (MSE) loss is applied:
\begin{equation}\small
\label{eq:mse_loss}
\begin{aligned}
     \mathcal{L} = \mathbb{E}_{\bm{x}_{0},\bm{\varepsilon},t}(\left \| \bm{\varepsilon}_{\theta}(\bm{x}_{t}, t) - (\bm{x}_{1}-\bm{x}_{0})  \right \|^{2}).
\end{aligned}
\end{equation}
This framework ensures accurate denoising, gradually recovering the original data from noisy latents.

\subsection{Implementation and Dataset Details}
We train the model using AdamW ($\beta_{1}=0.9$, $\beta_{2}=0.999$) and run the entire optimization in bfloat16. Distributed data parallelism is handled through DeepSpeed-Stage-3, which manages gradient synchronization and memory efficiency during training.

In terms of the training dataset, our training dataset consists of three parts, including Hallo3~\cite{cui2025hallo3}, Celebv-HQ~\cite{zhu2022celebvhq}, and collected videos from the internet (BilBil, YouTube, and TikTok).
We utilize the Q-Align~\cite{wu2023q} to filter for higher-quality videos by assessing the overall video fidelity. We also apply InsightFace~\cite{deng2019arcface} to filter out videos with a facial confidence score below 0.8. We obtain the final training dataset, containing roughly 2000 hours of videos.

\begin{figure}[t!]
\begin{center}
\includegraphics[width=0.8\linewidth]{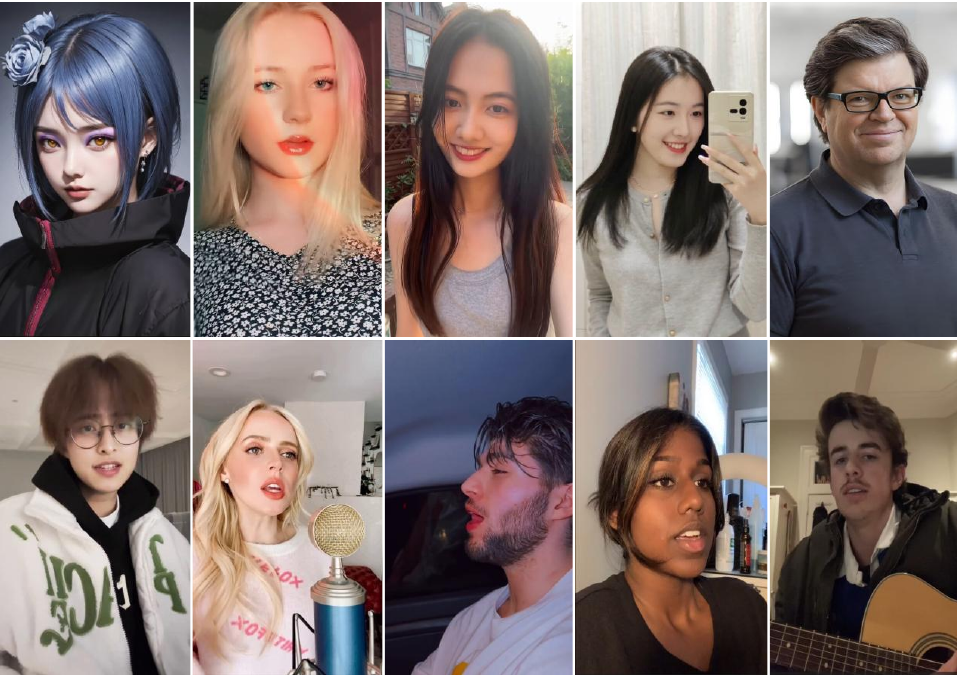}
\end{center}
\vspace{-0.5cm}
   \caption{Examples from Hard100.
   }
\label{fig:hard100}
\vspace{-0.3cm}
\end{figure}

Regarding the testing dataset, we first randomly select 100 videos (5-20 seconds long) from Voxceleb2~\cite{chung2018voxceleb2} and Vfhq~\cite{xie2022vfhq} to construct the first simple testing dataset. To validate the robustness of our FlashPortrait, we further select 100 unseen videos (1-3 minutes long, FPS=30) from the internet to construct the testing dataset Hard100. Some examples are shown in Fig. \ref{fig:hard100}. The sources of Hard100 come from various social media platforms, such as BilBil, YouTube, and TikTok. 
The selected videos span both indoor and outdoor environments, and the protagonists exhibit substantial demographic diversity, including balanced distributions across gender and ethnicity. The videos contain both upper-body and full-body subjects, with actions ranging from simple standing poses to complex interactions with objects in the scene. Consequently, our curated testing dataset is substantially more challenging than existing open-source testing datasets (Voxceleb2~\cite{chung2018voxceleb2} and Vfhq~\cite{xie2022vfhq}) in terms of subject diversity, environmental diversity, and pose variability. Moreover, the average duration of our selected videos is approximately two minutes, which is significantly longer than that of existing open-source testing datasets. Thus, it is more suitable for evaluating long-video generation performance.

\begin{table}[t!]\small
\caption{Ablation study on different weight assignment.
}
\vspace{-0.25in}
\begin{center}
\renewcommand\arraystretch{1.1}
\scalebox{0.85}{
\begin{tabular}{l|cccc}
\toprule
Model & AED$\downarrow$            & APD$\downarrow$            & MAE$\downarrow$            & Speed$\downarrow$         \\ \midrule
\textit{w/o} $\bm{s}(t)$ & 34.73          & 28.44          & 16.12          & 709s          \\
\textit{w/o} $\bm{w}(t, l, i)$ & 40.48          & 33.52          & 18.75          & \textbf{688s}          \\ \midrule
Ours  & \textbf{29.68} & \textbf{24.40} & \textbf{12.54} & 720s \\ \bottomrule
\end{tabular}
}
\end{center}
\label{table:more_acceleration}
\vspace{-0.4in}
\end{table}

\subsection{Additional Ablation on Acceleration}
We conduct an ablation study on two dynamic functions in our proposed Adaptive Latent Prediction Acceleration Mechanism, as shown in Table \ref{table:more_acceleration}. We observe that removing $\bm{s}(t)$ and $\bm{w}(t, l, i)$ significantly degrades performance.
It indicates that $\bm{s}(t)$ and $\bm{w}(t, l, i)$ can facilitate the accuracy of predicted latents based on the latent variation rate at particular timesteps and the derivative magnitude ratio among diffusion layers.
The underlying reason is that $\bm{s}(t)$ and $\bm{w}(t, l, i)$ jointly regulate the approximation between $\bigtriangleup^{i}\bm{f}(t, l)$ and $\bm{f}^{(i)}(t, l)$, ensuring robustness of latent prediction across diverse scenarios, even when the generated videos exhibit large motion variations.

\begin{figure*}[t!]
\begin{center}
\includegraphics[width=1\linewidth]{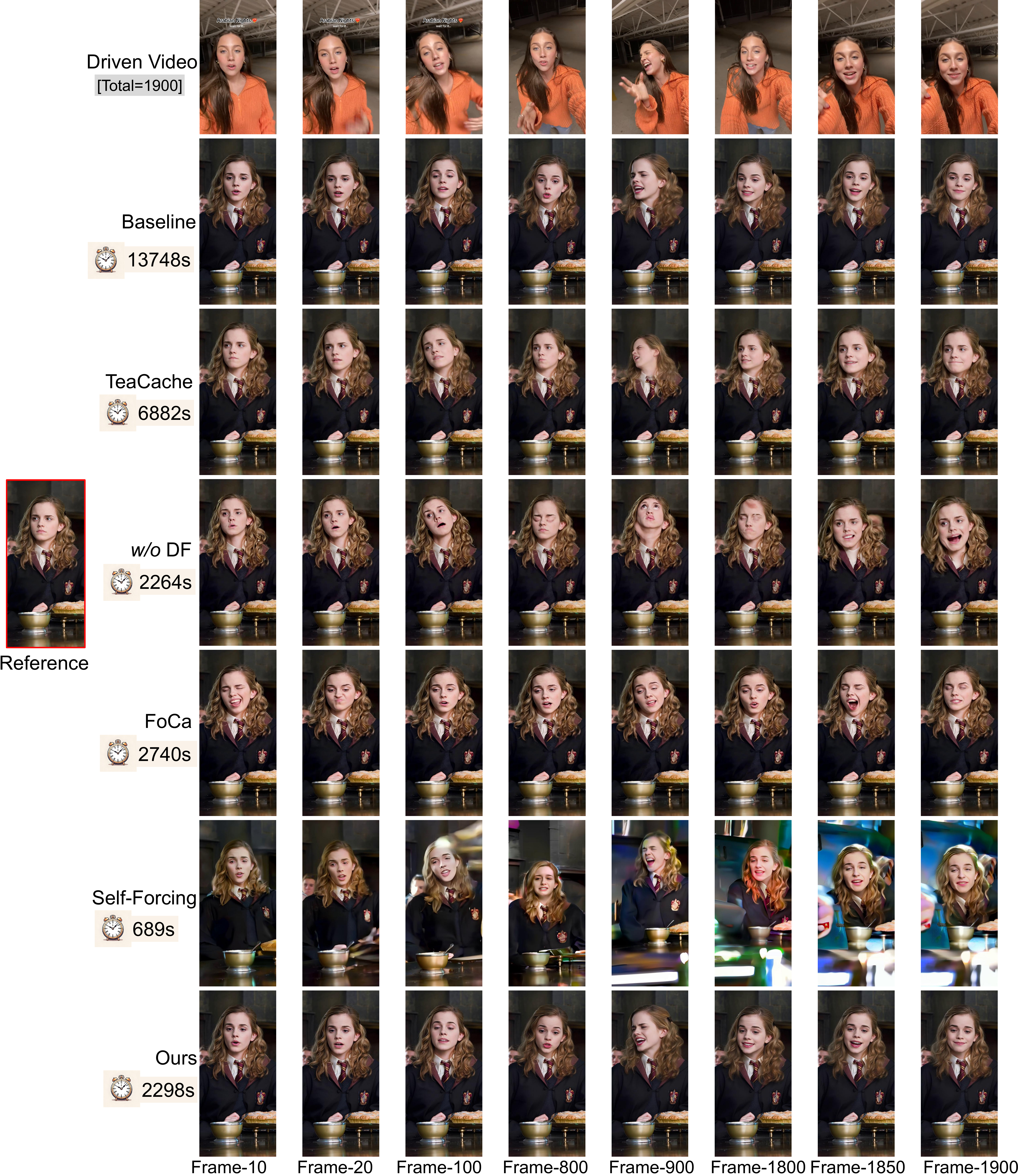}
\end{center}
\vspace{-0.55cm}
   \caption{Ablation study on different acceleration methods. \textit{w/o} DF refers to \textit{w/o} Dynamic Functions.}
\label{fig:more_acceleration}
\vspace{-0.5cm}
\end{figure*}

We further conduct an ablation study on different acceleration methods, presenting the results through progressive visualizations, as shown in Fig.\ref{fig:more_acceleration}. 
We observe that as the number of generated frames increases, all competitors become progressively unstable, particularly in terms of facial and background consistency. When the sequence length exceeds 800 frames, all competitors exhibit varying degrees of face and body distortion, as well as color drift. Moreover, the generated portrait no longer strictly follows the driven video, with facial expressions turning stochastic, especially in mouth closure, eye motion, and head rotation. By contrast, our FlashPortrait achieves a $6\times$ inference speedup over the baseline while maintaining comparable visual quality and preserving high-fidelity identity consistency. Moreover, the generated facial expressions strictly follow the guidance of the driven video, which demonstrates the superiority of our Adaptive Latent Prediction Acceleration Mechanism over previous acceleration methods in the long-length portrait animation.

\subsection{Full/Half Body Portrait Animation}
\begin{figure*}[t!]
\begin{center}
\includegraphics[width=1\linewidth]{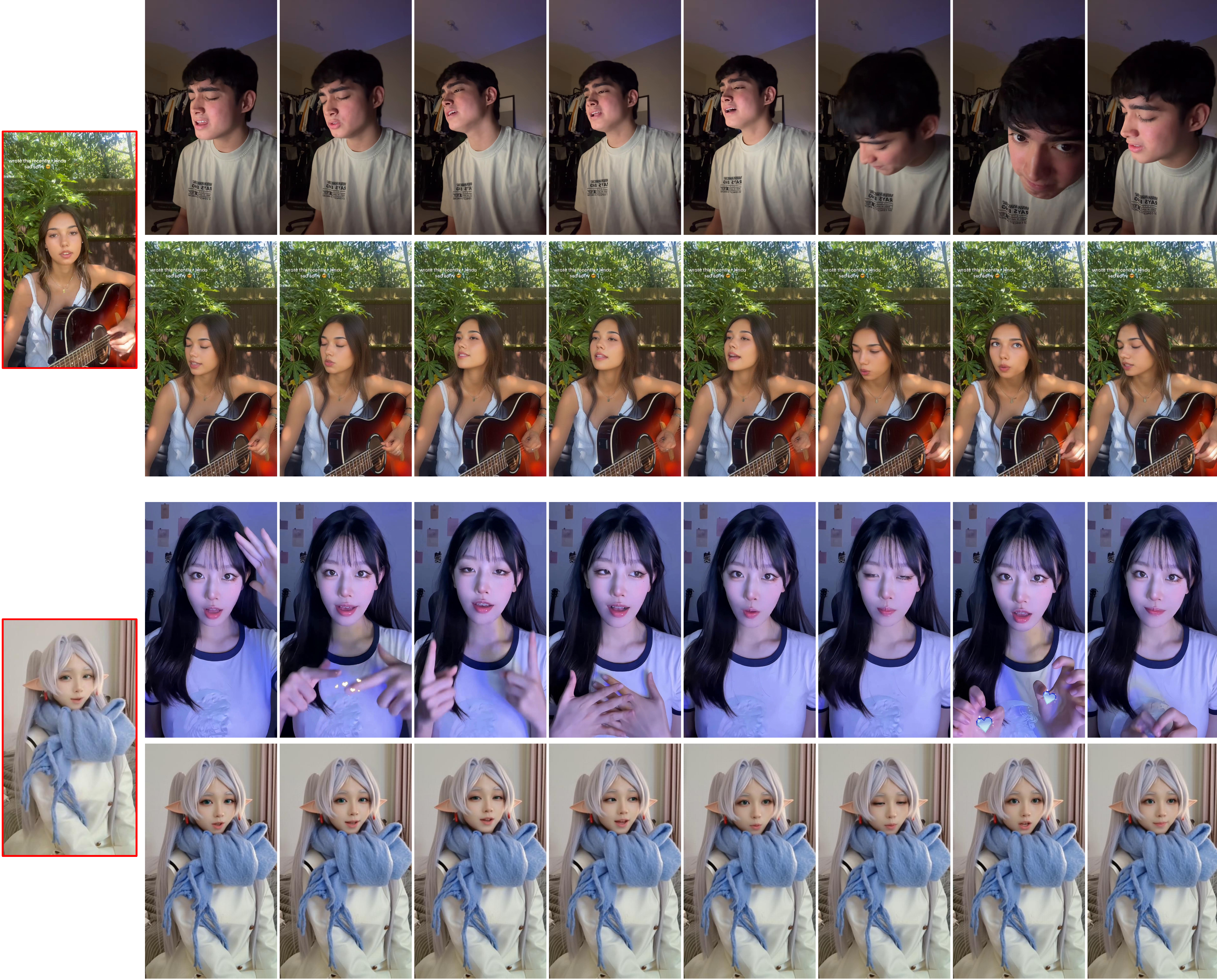}
\end{center}
\vspace{-0.55cm}
   \caption{Full/Half-body portrait animation results. The images with red borders are the reference images.}
\label{fig:full_body_video}
\vspace{-0.35cm}
\end{figure*}

We perform a qualitative experiment in full/half-body portrait animations, as shown in Fig. \ref{fig:full_body_video}.
Each reference image has a complex background layout and intricate foreground appearance. The first case even involves interactions with objects from the environment, such as an instrument, making it more challenging to maintain identity consistency and facial expression synchronization with the driven video.
We can see that our FlashPortrait has the capacity to synthesize full/half-body portrait animations, even involving interactions with external objects.

\section{Long Portrait Animation}
\begin{figure*}[t!]
\begin{center}
\includegraphics[width=1\linewidth]{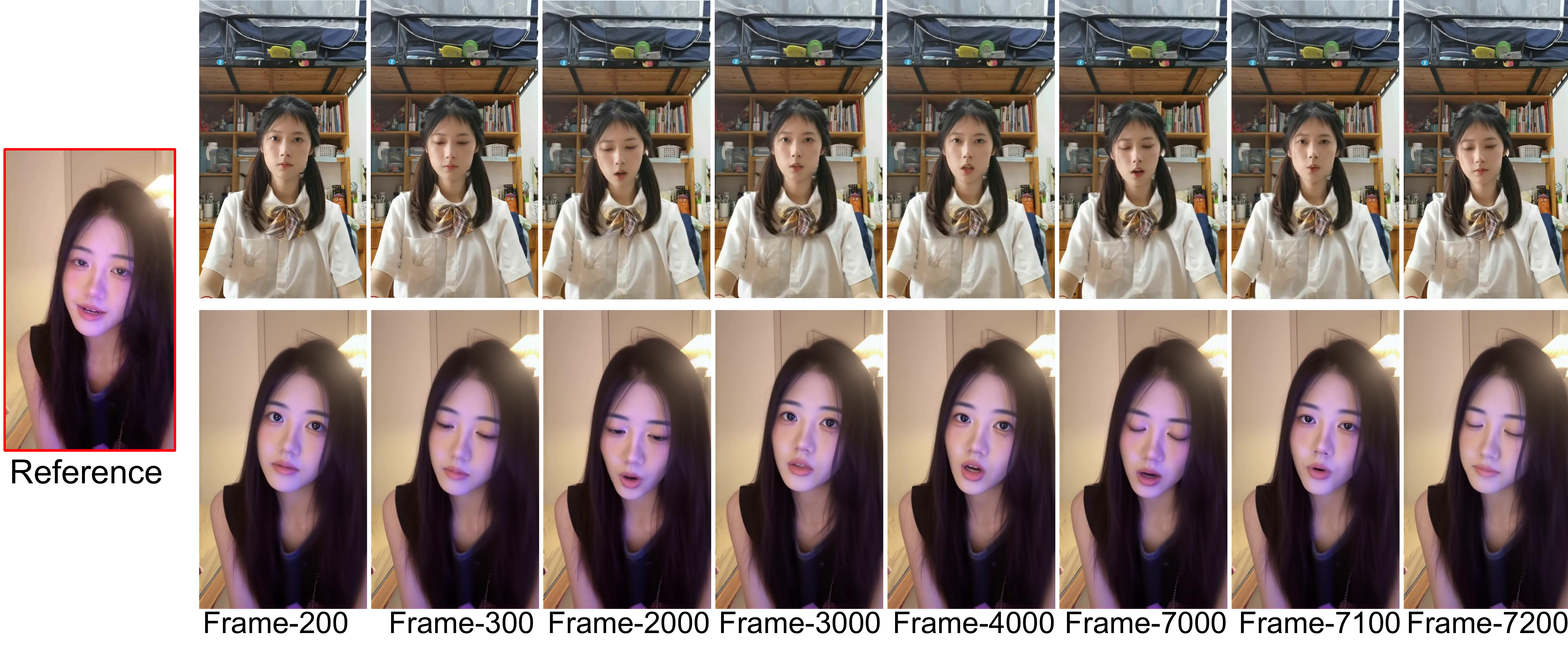}
\end{center}
\vspace{-0.55cm}
   \caption{Long portrait animation results. The images with red borders are the reference images.}
\label{fig:extreme_long_video}
\vspace{-0.5cm}
\end{figure*}

To further validate the performance of our FlashPortrait in long-length portrait animation, we perform a qualitative experiment in an extremely long case (4 minutes, FPS=30), as shown in Fig.\ref{fig:extreme_long_video}. Our FlashPortrait can still maintain identity consistency and ensure expression synchronization with the driven video, even after synthesizing 7000+ frames.
From a theoretical perspective, FlashPortrait can synthesize infinite-length high-quality identity-preserving animations.

\subsection{More Portrait Animation}
\begin{figure*}[t!]
\begin{center}
\includegraphics[width=0.85\linewidth]{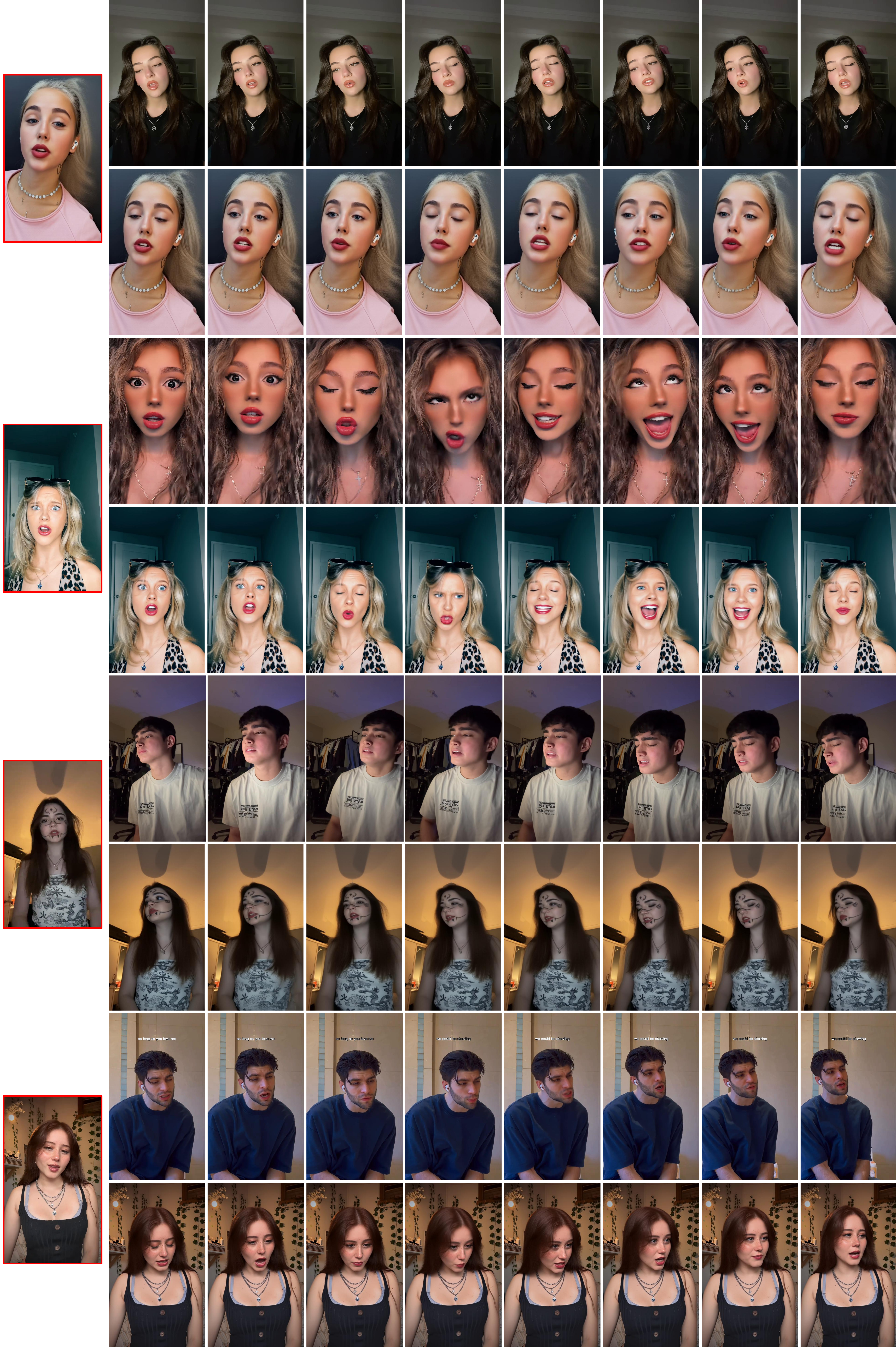}
\end{center}
\vspace{-0.55cm}
   \caption{portrait animation results (1/5). The images with red borders are the reference images.}
\label{fig:more_video_1}
\vspace{-0.5cm}
\end{figure*}

\begin{figure*}[t!]
\begin{center}
\includegraphics[width=0.85\linewidth]{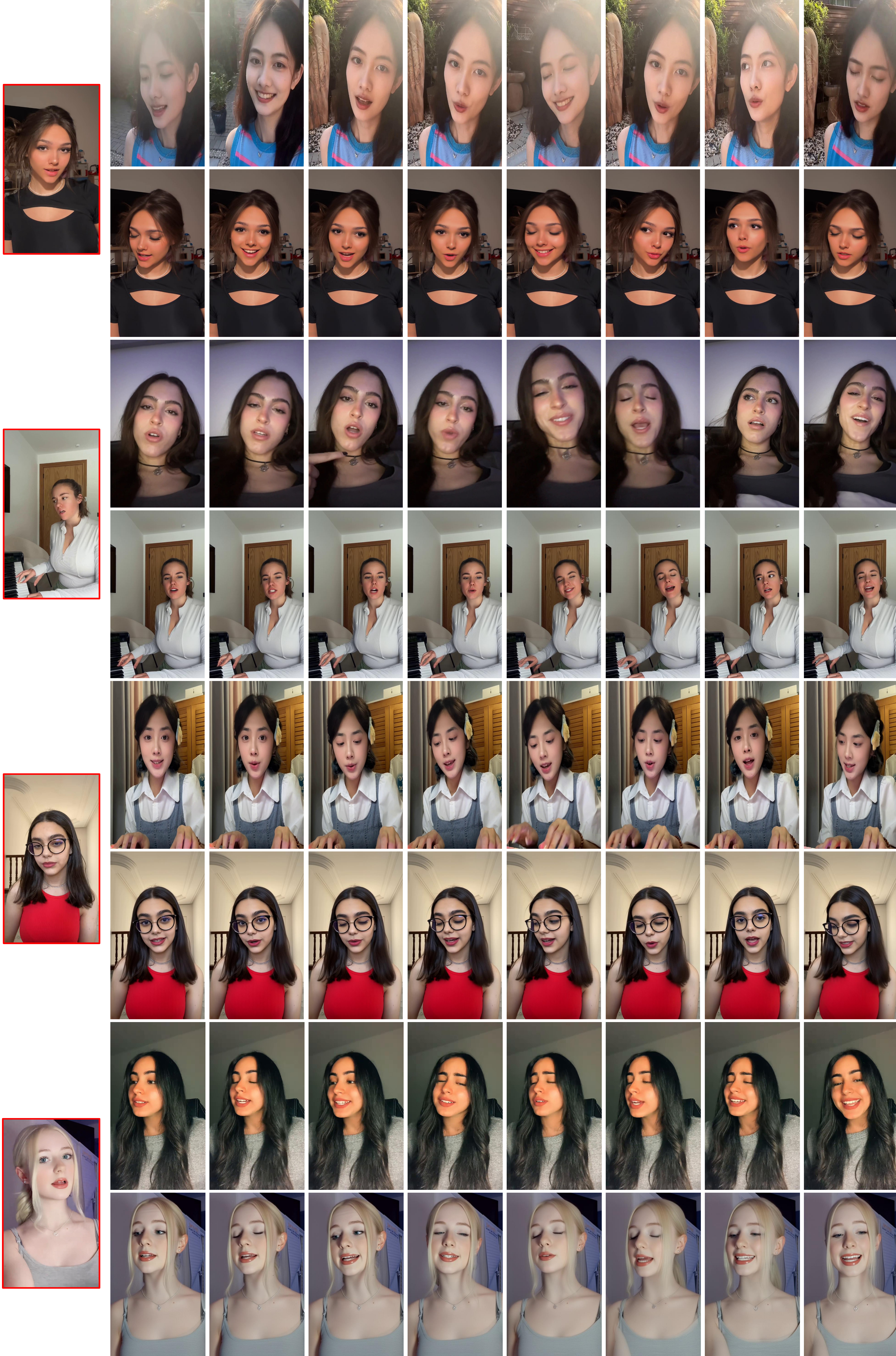}
\end{center}
\vspace{-0.55cm}
   \caption{portrait animation results (2/5). The images with red borders are the reference images.}
\label{fig:more_video_2}
\vspace{-0.5cm}
\end{figure*}

\begin{figure*}[t!]
\begin{center}
\includegraphics[width=0.85\linewidth]{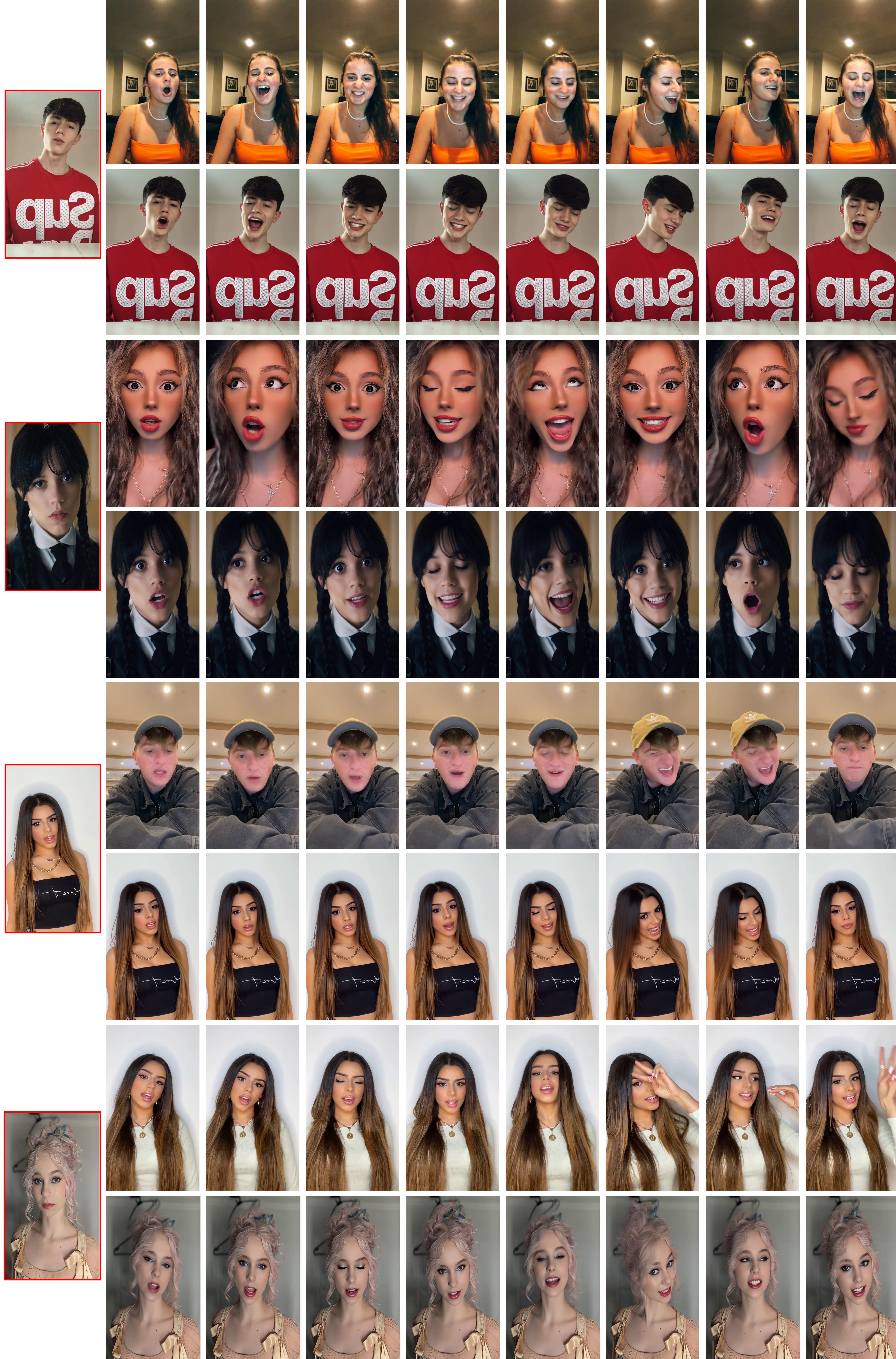}
\end{center}
\vspace{-0.55cm}
   \caption{portrait animation results (3/5). The images with red borders are the reference images.}
\label{fig:more_video_3}
\vspace{-0.5cm}
\end{figure*}

\begin{figure*}[t!]
\begin{center}
\includegraphics[width=0.85\linewidth]{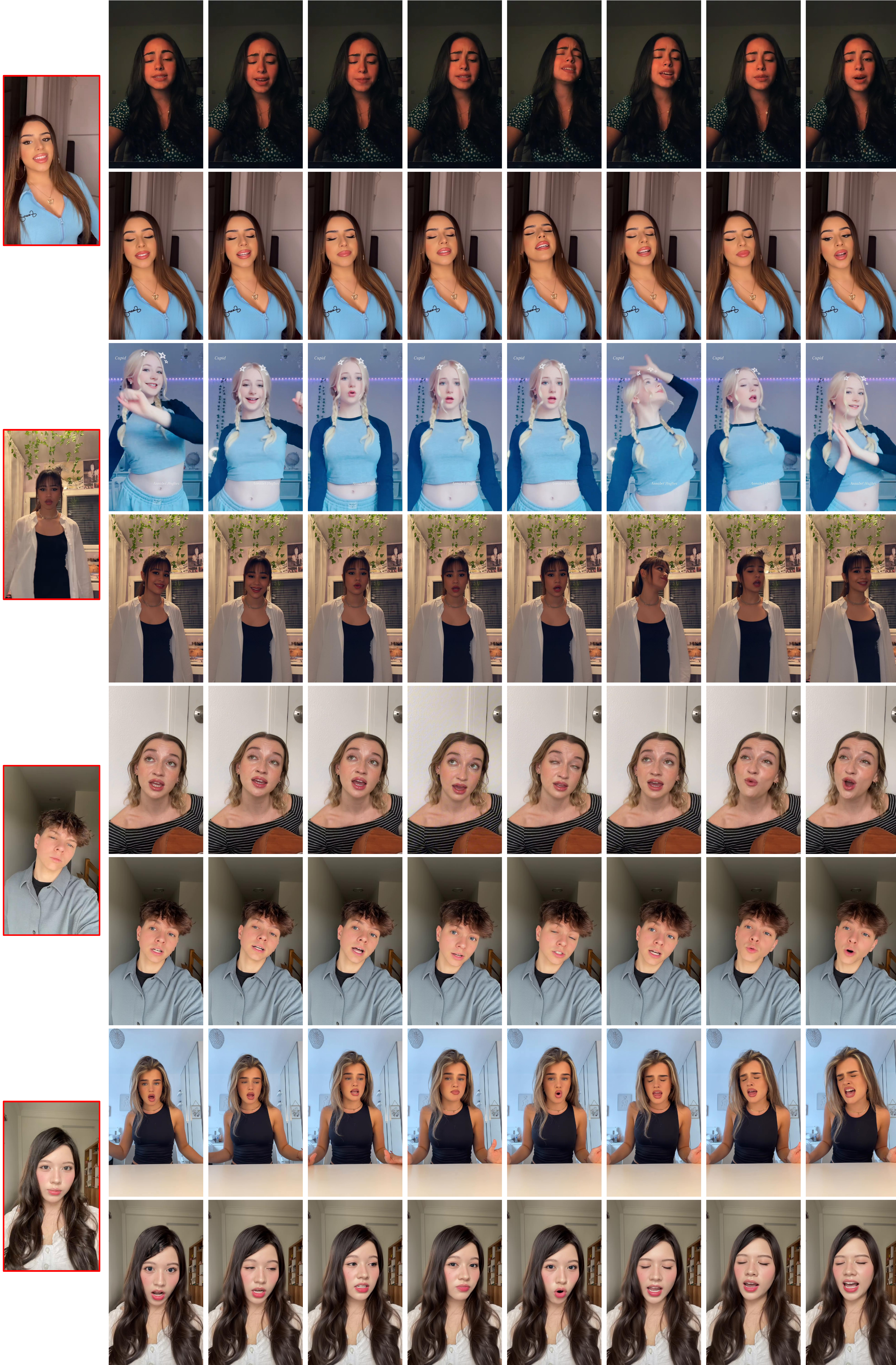}
\end{center}
\vspace{-0.55cm}
   \caption{portrait animation results (4/5). The images with red borders are the reference images.}
\label{fig:more_video_4}
\vspace{-0.5cm}
\end{figure*}

\begin{figure*}[t!]
\begin{center}
\includegraphics[width=0.85\linewidth]{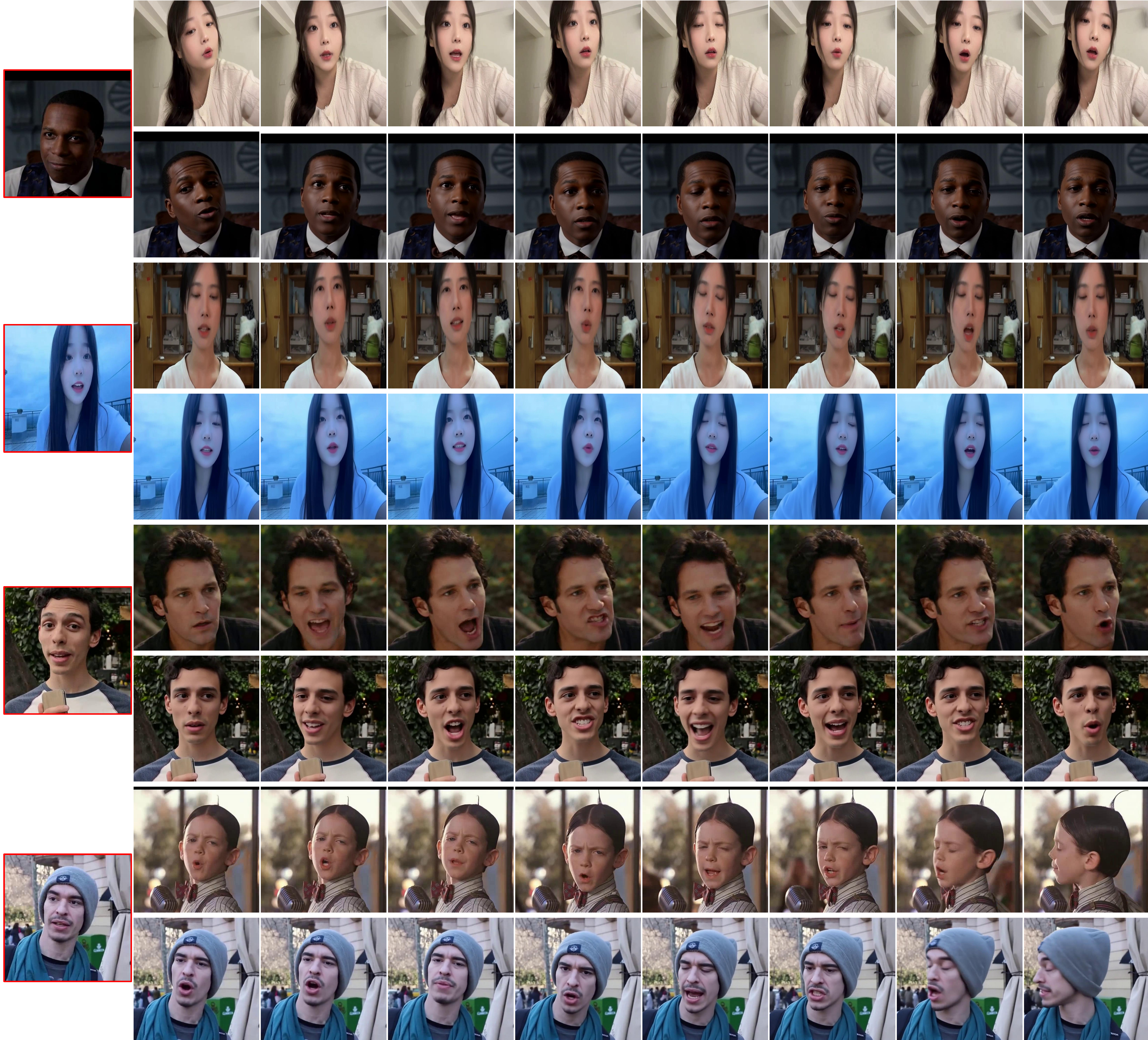}
\end{center}
\vspace{-0.55cm}
   \caption{portrait animation results (5/5). The images with red borders are the reference images.}
\label{fig:more_video_5}
\vspace{-0.5cm}
\end{figure*}

Fig. \ref{fig:more_video_1}, Fig. \ref{fig:more_video_2}, Fig. \ref{fig:more_video_3}, Fig. \ref{fig:more_video_4}, and Fig. \ref{fig:more_video_5} presents additional portrait animation result synthesized by our FlashPortrait. 
Each driven video contains 1800+ frames, and we only select synthesized frames from the last 100 frames for presentation. 
The reference protagonists exhibit rich diversity, encompassing both male and female subjects across various ethnicities. They also present complex visual characteristics, including intricate hairstyles, richly textured clothing, elaborate tattoo patterns, and a wide range of refined accessories. Each driven video contains substantial and dynamic facial expression motions with irregular expression patterns, such as head rotations and rapid blinking.
We can observe that our FlashPortrait can accurately animate the reference image based on the driven video while maintaining strong identity consistency even after synthesizing 1800 frames.
For example, the third row of Fig. \ref{fig:more_video_3} contains dramatic facial expression motions and exaggerated expression patterns, making it challenging for portrait animation model to preserve identity consistency while following the guidance of the driven video. 
Our FlashPortrait can still accurately manipulate the facial expression of the reference image (lip movement, eye movement, head movement) while maintaining high-quality identity consistency.

\subsection{Limitation and Future Work}
\begin{figure*}[t!]
\begin{center}
\includegraphics[width=1\linewidth]{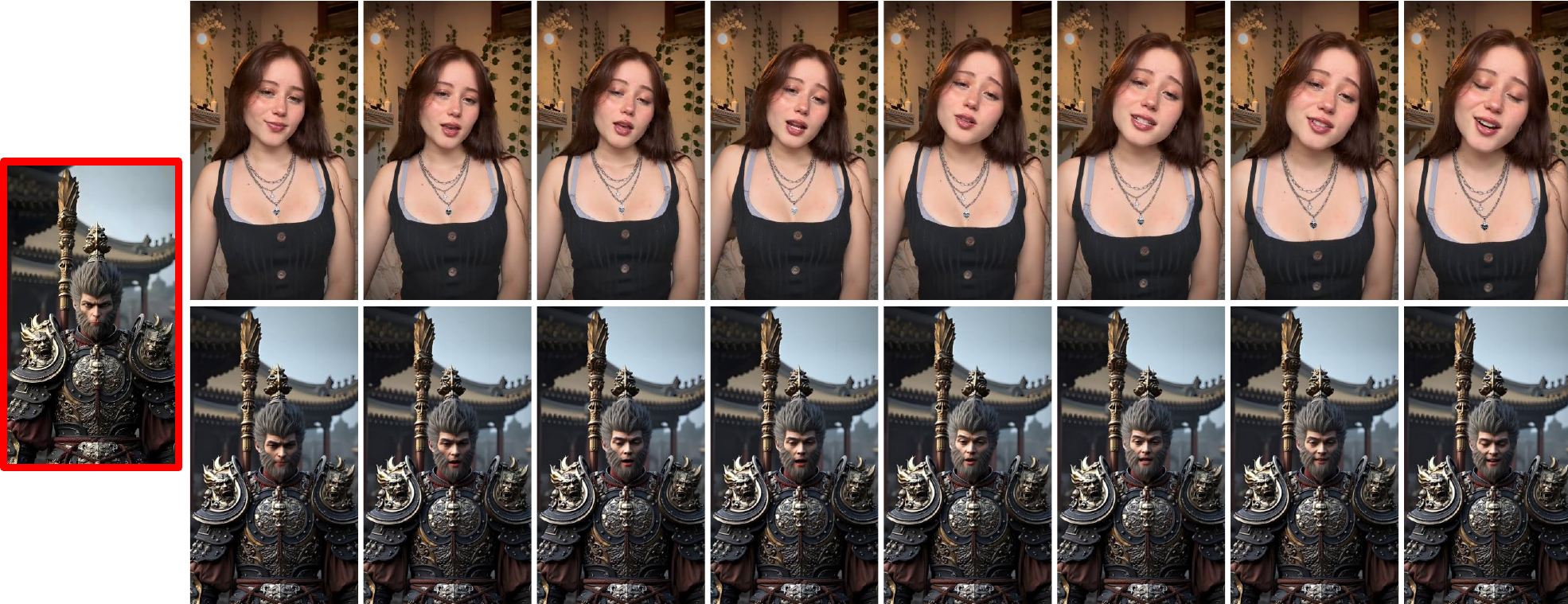}
\end{center}
\vspace{-0.55cm}
   \caption{One failure case of our FlashPortrait. The images with red borders are the reference images.}
\label{fig:limitation}
\vspace{-0.5cm}
\end{figure*}

Fig. \ref{fig:limitation} shows one failure case of our FlashPortrait.
When the reference protagonist is a humanoid character, such as a game avatar or a mythological figure, its appearance does not strictly conform to real human facial standards. Since our model is primarily trained on real human video data, FlashPortrait tends to synthesize a more realistic human face to replace the original reference protagonist’s face. This adaptation disrupts identity consistency and results in generated faces that deviate substantially from the reference image.
One potential solution is to introduce an additional reference network to explicitly capture the face details of the reference images. 
This reference network needs to be trained from scratch on large-scale diverse video datasets.
This part is left as future work. 

\subsection{Ethical Concern}
Our FlashPortrait can animate the reference image based on the driven video, allowing a reference image to be dynamically reenacted according to a driven video. This capability presents a risk of being exploited for deceptive media synthesis on social platforms. To address this risk, the deployment of robust sensitive-content and misuse-detection mechanisms is necessary to ensure responsible usage.

\end{document}